\documentclass[lettersize,journal]{IEEEtran}
\usepackage{amsmath,amsfonts}
\usepackage{algorithmic}
\usepackage{array}
\usepackage[caption=false,font=normalsize,labelfont=sf,textfont=sf]{subfig}
\usepackage{textcomp}
\usepackage{stfloats}
\usepackage{url}
\usepackage{verbatim}
\usepackage{graphicx}
\usepackage{cite}
\usepackage{booktabs}
\usepackage{CJKutf8}
\def\BibTeX{{\rm B\kern-.05em{\sc i\kern-.025em b}\kern-.08em
    T\kern-.1667em\lower.7ex\hbox{E}\kern-.125emX}}
\usepackage{balance}
\begin{document}
\begin{CJK}{UTF8}{gbsn}
\title{Local Map Construction with SDMap: A Comprehensive Survey}
\author{    
    Jiaqi Li$^{*}$, Pingfan Jia$^{*}$, Jiaxing Chen, Jiaxi Liu, Lei He$^{*}$, Keqiang Li
    \thanks{Jiaqi Li is with the Department of Civil Engineering, Tsinghua University, Beijing, 100084, China (e-mail: lijq22@mails.tsinghua.edu.cn). 
    Pingfan Jia, a co-first author, is with the School of Computer Science, Beihang University, Beijing, China (e-mail: pingfan@buaa.edu.cn).
    Jiaxing Chen is with the School of Vehicle and Mobility, Tsinghua University, Beijing, 100084, China (e-mail: chenjx23@mails.tsinghua.edu.cn). 
    Jiaxi Liu is with the Civil and Environmental Engineering Department, University of Wisconsin-Madison, WI, 53715, USA (e-mail: jliu2487@wisc.edu). Lei He and Keqiang Li is with the School of Vehicle and Mobility, Tsinghua University, Beijing, 100084, China. Lei He is the corresponding author (e-mail: helei2023@mail.tsinghua.edu.cn).}
}

\markboth{Journal of \LaTeX\ Class Files,~Vol.~18, No.~9, September~2020}%
{How to Use the IEEEtran \LaTeX \ Templates}

\maketitle

\begin{abstract}
Local map construction is a vital component of intelligent driving perception, offering necessary reference for vehicle positioning and planning. Standard Definition map (SDMap), known for its low cost, accessibility, and versatility, has significant potential as prior information for local map perception. This paper mainly reviews the local map construction methods with SDMap, including definitions, general processing flow, and datasets. Besides, this paper analyzes multimodal data representation and fusion methods in SDMap-based local map construction. This paper also discusses key challenges and future directions, such as optimizing SDMap processing, enhancing spatial alignment with real-time data, and incorporating richer environmental information. At last, the review looks forward to future research focusing on enhancing road topology inference and multimodal data fusion to improve the robustness and scalability of local map perception.
\end{abstract}


\section{Introduction}
\IEEEPARstart{I}{n} intelligent driving system, the construction of a local map is crucial for providing real-time awareness of the vehicle's surroundings, enabling precise navigation. It reflects the current state of the environment using sensor data and helps in predicting the movement of surrounding objects. By offering real-time updates, local maps support dynamic decision-making and control, even in unmapped areas. They enhance robustness, reduce computational load, and ensure that the vehicle can safely adapt to rapidly changing conditions. Local maps provide accurate static information such as road geometries and traffic sign locations, serving as prior knowledge that enhances the robustness of perception systems in complex environments. Local map perception systems are expected to achieve high-precision detection in all weather and lighting conditions, as any instance of missed or incorrect detection can result in significant consequences.

High-definition map (HDMap) is a vital topic in previous research. HDMap provides high precision, freshness, and richness of electronic map. However, HDMap faces significant challenges, primarily in terms of real-time updates and cost control. Urban road environments change frequently, and any minor alteration can impact the driving safety of autonomous vehicles. Traditional HDMap production methods require substantial time and resources, making real-time updates difficult. By heavily relying on perception and reducing dependence on detailed maps, this approach has gained widespread recognition. It emphasizes the use of onboard sensors for intelligent driving perception tasks, supplemented by lightweight map information. This strategy reduces reliance on real-time map updates, lowering maintenance costs, while lightweight map information can effectively compensate for certain limitations of onboard sensors, enhancing the model's robustness. SDMap, as a widely used electronic map in traffic navigation and geographic information services, features low production and maintenance costs, easy accessibility, and small data size, making it suitable as lightweight map to assist onboard sensors in constructing local map for intelligent driving.

\begin{figure*}[t!]
\begin{center}
\includegraphics[width=0.95\linewidth]{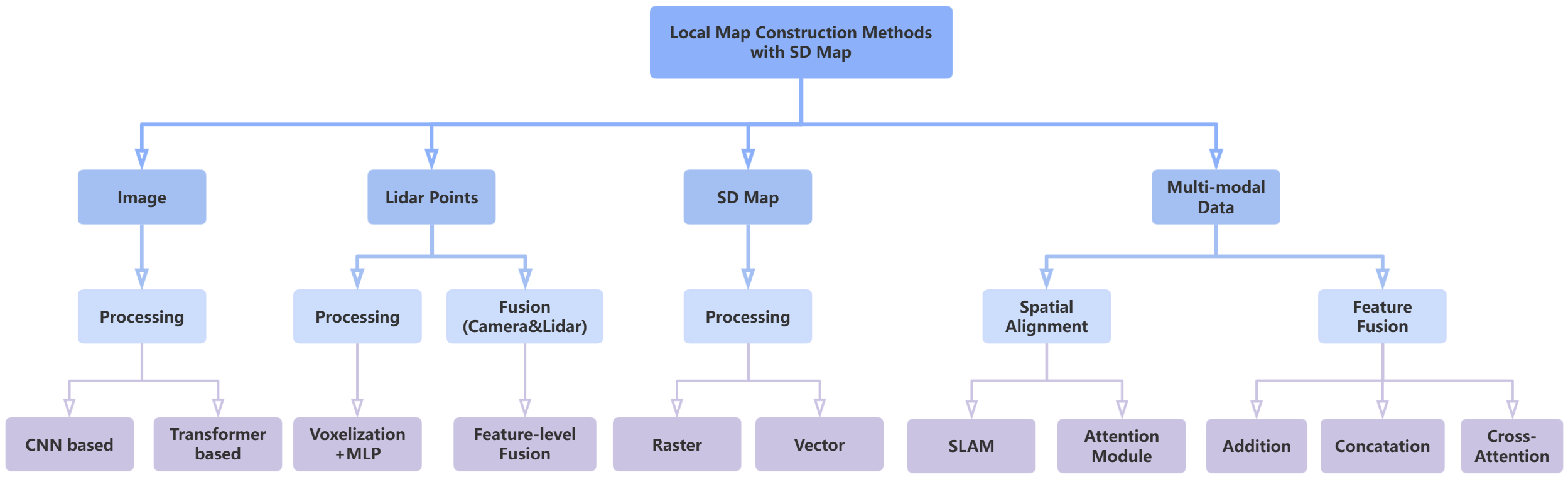}

\end{center}
   \caption{The overview of local map construction with SDMap.}
\label{overview}
\end{figure*}

Despite the promising prospects and numerous challenges in constructing local map based on SDMap, there is a lack of comprehensive research reviews in this area. To address this gap, this review aims to provide a thorough overview of the latest advancements in local map construction methods utilizing SDMap. Specifically, the focus is on the application of SDMap information representation methods and multimodal data fusion techniques in local map construction tasks. This research delves into the major developments, challenges, and research directions in this field. The contributions to the existing body of knowledge are as follows, The existing literature on local map construction using SDMap as a prior is reviewed. The representation and encoding methods of information from various sensors, as well as the fusion techniques for multimodal data. The underlying principles, architecture, and performance of these methods are discussed, shedding light on their feasibility and practicality in the field key challenges and open research questions in local map construction using SDMap as a prior are identified.

Fig. \ref{overview} presents the framework and context of the comprehensive review, where the relationships between the modules in the figure are organized in order and hierarchy. This paper analyzes and organizes from the perspective of processing multimodal data. The data is categorized into images, lidar point clouds, SDMap data, and multimodal data. Each category corresponds to specific processing and fusion methods, with distinct techniques and formats for handling and integration. It can be seen in the figure that in the local map construction method with SDMap, the image processing methods are divided into CNN-based modules or transformer-based modules, while the point cloud data is divided into extraction processing methods and fusion methods with image data. The processing methods for SDMap data are divided into Raster and Vector categories based on the form of processed data. Finally, the processing methods for multimodal data are divided into spatial alignment and feature fusion, the former is further divided into SLAM-based and attention-based methods, while the latter fusion methods are divided into addition, concatenation, and cross-attention methods. This classification structure helps to clearly sort out the different processing means and their characteristics in local map construction methods with SDMap.

Following this introduction, the article unfolds as follows: Section \ref{sec:related_work} provides a comprehensive review of the pertinent literature, elucidating the general pipeline for local map construction utilizing SDMap. Section \ref{sec:multi} delves into the concept of SDMap, its advantages, and its application scenarios within the task of local map construction. Additionally, we present an overview of the public datasets commonly employed and the evaluation metrics pertinent to tasks of local map perception. Section \ref{sec:mulfusion} scrutinizes the representation, encoding, and fusion methods of multi-source information, which are crucial for the generation of real-time local maps. This paper further discusses the underlying principles, architecture, and performance of these methods. Section \ref{sec:challenges} identifies and discusses key challenges and potential research directions associated with leveraging SDMap as prior information for local map construction. Finally, Section \ref{sec:conclusion} synthesizes our findings and offers perspectives on future research endeavors.

\begin{figure*}
\begin{center}
\includegraphics[width=0.95\linewidth]{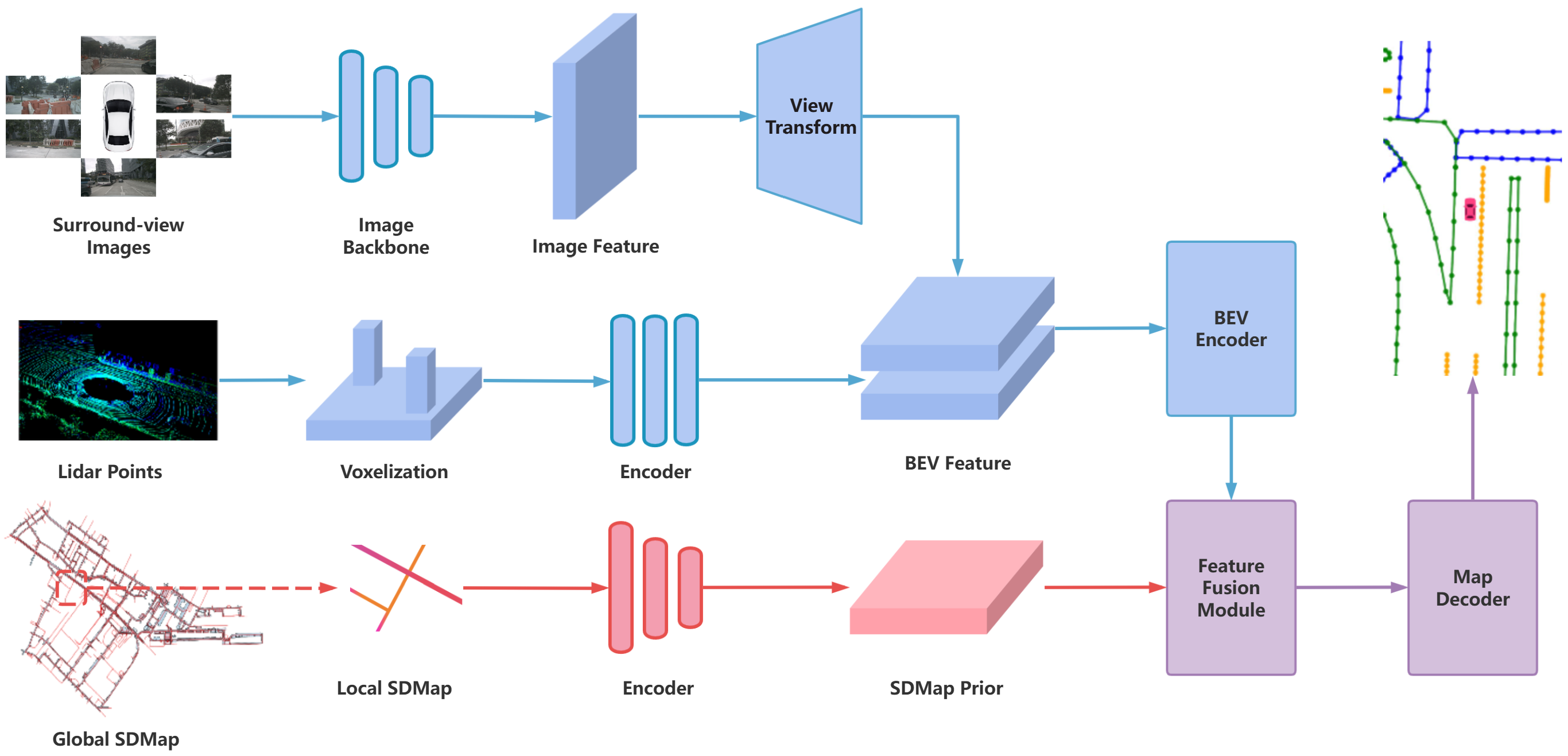}

\end{center}
   \caption{General pipeline of local map construction with SDMap, which contains the processing and fusion flow of the three modal data.}
\label{fig:short}
\end{figure*}

\section{Related Work} \label{sec:related_work}
\noindent Sensor noise and data delays further complicate the perception task. Therefore, developing robust local map perception technologies is crucial for achieving safe and reliable intelligent driving.
To solve this kind of problem, many researchers have proposed various methods. Chen and et al.\cite{VLMCGTex} proposed a method for visual localization and map construction using ground texture, enhancing the accuracy of positioning and the precision of map updates through global and local optimization. SMERF algorithm\cite{SMERF} enhances online map prediction and lane topology understanding by leveraging SDMap and integrating SDMap information through a transformer encoder, which alleviates problems with obscured lane lines or poor visibility and significantly improves the performance of lane detection and topology prediction. \cite{OMR} proposed an innovative video lane detection algorithm that enhances the feature map of the current frame using an occlusion-aware memory-based refinement (OMR) module, leveraging obstacle masks and memory information to improve the detection accuracy and robustness under occlusion. RVLD\cite{RVLD} improved the reliability of lane detection by recursively propagating the state of the current frame to the next frame, utilizing the information from the previous frames. Besides, there are LaneAF\cite{Laneaf}, LaneATT\cite{LaneATT}, and StreamMapNet\cite{Streammapnet} to ease these issues.

In this section, the definition of local map construction with SDMap is clarified, and the general pipeline for this type of task is summarized. The composition and application scenarios of SDMap are introduced. Finally, commonly used public datasets and evaluation metrics in local map perception tasks are listed.

\subsection{The Task Definition}

The task of local map perception involves creating an accurate map representing the vehicle's surrounding environment to support intelligent driving decision-making and planning. This task typically relies on data from various sensors, including cameras, lidar, radar, and GPS. Additionally, incorporating prior information from SDMap enhances the model's robustness and mitigates the impact of uncertainties from onboard sensors, improving the overall model performance. The core of the local map perception task is real-time sensing and understanding of the vehicle's surroundings.

The general process of neural networks used for local map construction can be summarized into several key components, as illustrated in Fig. \ref{fig:short}. After inputting surround view images and lidar point clouds, the overall architecture of the local map construction network can be depicted as consisting of different parts: a backbone for image feature extraction, a PV2BEV (Perspective View to Bird's Eye View) module for perspective transformation, a module for multimodal feature fusion, and task-specific heads for lane detection. These components form the basic framework of the local map perception network. The images and point cloud data captured by the surround view cameras and lidar are first processed by a backbone to obtain (multi-scale) image features. These features are then transformed to the BEV perspective using the PV2BEV module, followed by fusion with SDMap data through a modality fusion module, and finally output through different task-specific heads.

\subsection{Standard Definition Map}

SDMap, short for Standard Definition Map, is a digital map technology providing basic geographic information and road network structures. It is widely used in everyday navigation and geographic information services, offering convenience to users. SDMap primarily provides the centerline skeleton of roads without detailed lane information, road signs, or other high-precision environmental features. For the task of local map construction, SDMap offers three main advantages. First, SDMap data is easily accessible. It can typically be obtained for free from open geographic data sources such as OpenStreetMap (OSM)\cite{Openstreetmap}, making it suitable for large-scale applications. Second, compared to HDMap, the production and maintenance costs of SDMap are significantly lower. Lastly, SDMap has high universality, covering most types of roads, and can provide relevant road information for local map construction tasks. Platforms like OSM and Baidu map can serve as data sources for SDMap. For instance, OSM is a collaborative project created and maintained by global volunteers, providing free, editable, and open-content maps. OSM data includes a wide range of geographic information such as roads, buildings, parks, and rivers, which can be freely accessed.

\subsection{Datasets}

In the field of BEV local map construction, commonly used datasets include KITTI\cite{kitti}, nuScenes\cite{nuscenes}, ApolloScape\cite{apolloscape}, Argoverse\cite{Argoverse}, Openlane\cite{Openlane}, and Waymo\cite{Waymo} Open Dataset (Table \ref{tab:datasetTab}). 

\begin{table*}[t]
\begin{center}
\caption{Datasets for BEV local map construction} 
\label{tab:datasetTab} 
\begin{tabular}{| c | c | c | c | c | c | c | c | c | c |} 
\hline
\toprule 
dataset & Year & Locations & Scenes & Train & Val & Test & Views & Map Element & HDMap \\
\hline
KITTI\cite{kitti} & 2012 & Germany & 22	& - & - & - & 1 & - & -\\
\hline
Argoverse\cite{Argoverse} & 2019 & USA & 113 & 200k & 40k & 80k & 7 & Only lanes & √\\
\hline
nuScenes\cite{nuscenes}  & 2020 & USA/Singapore & 1000 & 700 & 150 & 150 & 6 & - & √\\
\hline
WaymoOpen\cite{Waymo} & 2020 & USA & 1150 & 798 & 202 & 150 & 5 & Only signs & -\\
\hline
CurveLanes\cite{Curvelane} & 2020 & China & - & 100k & 20k & 30k & 1 & - & -\\
\hline
ONEC-3DLanes\cite{Once-3dlane} & 2022 & China & - & 200k & 3000 & 8000 & 1 & - & -\\
\hline
OpenLane-V2\cite{Openlane-v2} & 2023 & USA/Singapore & 2000 & 1400 & 300 & 800 & 6/7 & √ & √\\
\hline
\end{tabular}
\end{center}
\end{table*}

The KITTI\cite{kitti} dataset, created by the Karlsruhe Institute of Technology and the Toyota Technological Institute, provides stereo camera, lidar, and GPS/IMU data, covering urban, rural, and highway scenes, and is suitable for tasks such as object detection, tracking, and road detection. The KITTI dataset offers 3D bounding box tracklets for dynamic objects within the camera's field of view. It defines various classes such as 'Car', 'Van', 'Truck', 'Pedestrian', and provides annotations for each object's 3D size, translation, rotation, and occlusion levels.

nuScenes\cite{nuscenes}, released by Motional, includes data from six cameras, five radars, one lidar, IMU, and GPS, suitable for urban traffic scenarios under various weather and lighting conditions. The nuScenes dataset\cite{nuscenes} features comprehensive annotations for intelligent driving, including 3D bounding boxes for 23 classes of objects and 8 attributes. Each scene, lasting 20 seconds, is fully annotated with detailed 3D bounding boxes. The annotation process involves sampling keyframes at 2Hz and is performed by expert annotators to ensure accuracy. The dataset additionally includes continuous annotations for objects across each scene, provided they are captured by lidar or radar points.

ApolloScape\cite{apolloscape}, released by Baidu, offers high-precision 3D annotated data covering various urban road scenes, suitable for tasks like lane detection and semantic segmentation. The ApolloScape dataset\cite{apolloscape} includes per-pixel semantic segmentation annotations with a unique scale and complexity. It features over 140K images with pixel-level semantic masks and is planned to scale up to 1 million images. The dataset is annotated using an interactive and efficient labeling pipeline that integrates 2D and 3D labeling stages, leveraging high-quality 3D point clouds to enhance the annotation process. Each image is accompanied by highly accurate pose information, and the dataset features detailed labeling of lane markings, making it the first publicly available dataset to offer 3D annotations for street views.

Argoverse\cite{Argoverse}, released by Argo AI, includes stereo camera, lidar, GPS, and IMU data, providing detailed 3D annotations and lane markings, primarily used for 3D object detection and lane detection. Argoverse\cite{Argoverse} provides detailed 3D tracking annotations, interesting vehicle trajectories, and rich semantic maps. The dataset covers 290 km of mapped lanes with geometric and semantic metadata, using map information to enhance the accuracy of 3D object tracking.

The Waymo Open Dataset\cite{Waymo}, released by Waymo, covers a variety of weather and traffic conditions, providing high-quality data from lidar and cameras, suitable for tasks such as 3D object detection, tracking, and lane detection. The Waymo\cite{Waymo} provides comprehensive 2D and 3D bounding box annotations for camera images and lidar data, respectively. The dataset consists of 1150 scenes, each 20 seconds long, with synchronized and calibrated data from urban and suburban areas. Annotations include consistent identifiers across frames to support object tracking.

OpenLane-V2\cite{Openlane-v2}, also known as OpenLane-Huawei or Road Genome, is a benchmark dataset for next-generation intelligent driving scene road structure perception, jointly open-sourced by Shanghai Artificial Intelligence Laboratory and Huawei Noah's Ark Laboratory. It is the first dataset to include the topological relationships of road structures in traffic scenes. OpenLane-V2\cite{Openlane-v2} focuses on topology reasoning for traffic scene structure, offering annotations that describe the relationship between traffic elements and lanes. The dataset comprises 2,000 annotated road scenes that detail traffic elements and their correlation to lanes, aiming to advance research in understanding the structure of road scenes.

ONCE-3DLanes\cite{Once-3dlane} dataset, a real-world intelligent driving dataset with lane layout annotation in 3D space, is a new benchmark constructed to stimulate the development of monocular 3D lane detection methods. It is collected in various geographical locations in China, including highways, bridges, tunnels, suburbs, and downtown, with different weather conditions (sunny / rainy) and lighting conditions (day / night). The whole dataset contains 211K images with its corresponding 3D lane annotations in the camera coordinates. 

CurveLanes\cite{Curvelane} is a new benchmark lane detection dataset with 150K lane images for difficult scenarios such as curves and multi-lanes in traffic lane detection. It is collected in real urban and highway scenarios in multiple cities in China. All images are carefully selected so that most of them image contains at least one curve lane. More difficult scenarios such as S-curves, Y-lanes, night and multi-lanes can be found in this dataset.

\subsection{Common Evaluation Metrics}

The review introduces the evaluation metrics for local map construction methods from two aspects: lane extraction and topology reasoning.

\subsubsection{Metrics for Lane Extraction}

Mean Average Precision (mAP) is a common metric used to evaluate the performance of object detection models. mAP measures the precision of a model at various threshold levels by matching predicted bounding boxes with ground truth boxes to calculate true positives (TP), false positives (FP), and false negatives (FN). Initially, predicted boxes are matched with ground truth boxes based on a specified IoU (Intersection over Union) threshold. Then, precision (TP / (TP + FP)) and recall (TP / (TP + FN)) are calculated for each class and used to plot the Precision-Recall curve. The area under this curve is calculated using interpolation methods to obtain the Average Precision (AP) for a single class. Finally, the mean of the AP values across all classes gives the mAP, reflecting the overall detection performance of the model, with higher values indicating better performance.
\begin{equation}
    mAP = \frac{1}{N} \sum_{i=1}^{N} \text{AP}_i \
\end{equation}

Mean Intersection over Union (mIoU) is a commonly used metric to evaluate the performance of semantic segmentation models. mIoU measures the classification accuracy of the model at the pixel level for various objects. The calculation involves several steps. For each class, the IoU is computed by dividing the number of intersecting pixels (Intersection) between the predicted and ground truth areas by the union of these areas (Union). This calculation is performed for each class, and the mean IoU across all classes gives the mIoU, providing an average performance evaluation of the model's segmentation accuracy, with higher values indicating better segmentation performance.
\begin{equation}
    mIoU = \frac{1}{C} \sum_{c=1}^{C} \frac{TP_c}{FP_c + FN_c + TP_c} \
\end{equation}

Traditional object detection metrics like mAP may not fully capture all important aspects of the detection task, such as the estimation of object speed and attributes, and the accuracy of position, size, and orientation. Therefore, the nuScenes\cite{nuscenes} Detection Score (NDS) has been proposed to comprehensively account for these factors. NDS integrates multiple key metrics to overcome the limitations of existing metrics and provide a more holistic performance evaluation.

The NDS calculation formula is as follows:
\begin{equation}
   NDS = \frac{1}{2} \times (\text{mAP} + \text{mATE})
\end{equation}

Where, mAP represents the mean Average Precision, measuring detection accuracy. The TP set contains the average values of five True Positive Metrics: ATE (Average Translation Error), ASE (Average Scale Error), AOE (Average Orientation Error), AVE (Average Velocity Error), and AAE (Average Attribute Error).

\subsubsection{Metrics for Topology Reasoning}

OpenLane-V2\cite{Openlane-v2} has two types of topological reasoning, namely driving scene topology and OpenLane topology. For driving scene topology, OpenLane-V2 defined the OpenLane-V2 UniScore (OLUS), which is the average of various metrics covering different aspects of the primary task. Where $DET_l$ represents mAP on lane segments, $DET_t$ represents mAP on $DET_l$ represents mAP on lane segments and $DET_a$ represents mAP on areas. Areas namely pedestrian crossings and road boundaries, are regarded as undirected curves, which are closed or intersected with the boundaries of the BEV range. Chamfer distance is utilized to describe the similarity of areas. $TOP_{ll}$ represents mAP on topology among lane segments and $TOP_{lt}$ represents mAP on topology between lane segments and traffic elements.
\begin{equation}
    \text{OLUS} = \frac{1}{5} \left[ \text{DET}_{l} + \text{DET}_{a} + \text{DET}_{t} + f(\text{TOP}_{ll}) + f(\text{TOP}_{lt}) \right]
\end{equation}

 For OpenLane topology, OpenLane-V2\cite{Openlane-v2} breaks down the task into three subtasks: 3D lane detection, traffic element recognition, and topology reasoning. The overall task performance is described using the OpenLane-V2 Score (OLS), which is the average of the metrics for each subtask. The metric for 3D lane detection, $DET_l$, can be expressed as the mean AP at different thresholds $ t\in T $, $ T $=\{1.0, 2.0, 3.0\}, where AP is calculated using the Fréchet distance. Traffic element detection is evaluated similarly to object detection using AP, with an IoU threshold set to 0.75. Traffic elements have various attributes, such as traffic light colors, which are closely related to lane accessibility, so attributes must also be considered. Assuming A is the set of all attributes, the evaluation includes attribute classification accuracy.
\begin{equation}
    \text{OLS} = \frac{1}{4} \left[ \text{DET}_{l} + \text{DET}_{t} + f(\text{TOP}_{ll}) + f(\text{TOP}_{lt}) \right]
\end{equation}
\begin{equation}
    DET_l = \frac{1}{|T|} \sum_{t \in T} \text{AP}_t \
\end{equation}
\begin{equation}
    DET_t = \frac{1}{|A|} \sum_{a \in A} \text{AP}_a \
\end{equation}

OpenLane-V2\cite{Openlane-v2} uses the TOP score to evaluate the quality of topology reasoning, akin to the mAP metric but adapted for graphs. Essentially, this converts the topology prediction problem into a link prediction problem and calculates mAP (mean APs of all vertices) to assess algorithm performance. The first step is to determine a matching method to pair ground truth and predicted vertices (i.e., centerlines and traffic elements). For centerlines, Fréchet distance is used; for traffic elements, IoU is used. When the confidence score of an edge between two vertices exceeds 0.5, they are considered connected. The vertex AP is obtained by ranking all predicted edges of a vertex and calculating the mean of cumulative precision:
\begin{equation}
    TOP = mAP = \frac{1}{|V|} \sum_{v \in V} \frac {\sum_{\hat n \in \hat N(v) }P(\hat n)1(\hat n \in N(v))}{|N(v)|} \
\end{equation}

\section{Multimodal Representation} \label{sec:multi}
\noindent In this section, our paper will explore the data modalities, including the extraction and processing of image data, point cloud data, and SDMap data, as well as an introduction to the fusion methods for the aforementioned three types of modal data.

\subsection{Image}
In the perception task of BEV, the image information of the panoramic camera is the most important input data, and the common feature extraction method of the panoramic image follows the paradigm of intelligent driving perception task BEVformer\cite{Bevformer} or Lift-Splat-Shoot (LSS)\cite{lss}. The backbone module of neural networks extracts 2D image features from various camera perspectives through classic and lightweight convolutional networks such as ResNet-50 or 101\cite{ResNet}, MobileNets\cite{Mobilenets}, EfficientNet\cite{efficientnet}, V2-99\cite{vov} and so on. Among them, the ResNet series\cite{ResNet} is widely used and variants like ResNet enhance feature extraction capabilities by increasing network depth and width. These networks are extensively utilized in BEV local map perception tasks due to their outstanding performance in image recognition and feature extraction. Typically, a Feature Pyramid Network (FPN)\cite{FPN} module is appended to the backbone module. The FPN\cite{FPN} integrates feature maps of different scales, generating more robust multi-scale feature representations. This seems to be the default basic configuration, and the number of fusion layers can be selected according to the network type. This multi-scale feature fusion aids in improving the detection and recognition of objects of varying sizes, thereby enhancing overall performance.

In addition to lightweight and simple backbone, larger backbone networks will be the mainstream trend in the future. With the success of transformers in the field of computer vision\cite{vit}, transformer-based feature extraction methods have also been applied to BEV local map perception tasks such as Swin Transformer\cite{Swin}. Refer to the methods on the nuScenes leaderboard, The state-of-the-art methods all use pre-trained VIT-L\cite{vit} as the backbone network, or its variant EVA-02\cite{eva}. This large pre-trained backbone network is the key to improving model performance, although the large number of parameters and computational complexity of large models can seriously affect inference speed. Nevertheless, its performance directly promotes detection accuracy.

The training of these large models requires massive data support, but data labeling is costly and limited, and self-supervised training methods will become mainstream. With the widespread application of BERT\cite{bert} pre-trained models in various downstream tasks for self-supervised tasks in natural language processing, it has demonstrated a powerful ability to learn language representations. Similarly, in self-supervised learning in computer vision tasks, MAE\cite{mae} randomly masks patches on images and implements self-supervised learning of masked images. The achievements of MIM\cite{mim} based pre-training algorithms are flourishing in the field of computer vision. This self-supervised pre-training model can not only solve the problem of high-cost labels, but also better learn the representation relationship of images.

Whether based on CNN or transformer methods, the ultimate goal is to obtain high-quality feature representations of panoramic images. For BEV local map perception tasks, feature representation is crucial as it directly affects the accuracy and robustness of the perception system. The global feature extraction mechanism of FPN module or transformer can significantly improve the overall performance of the network, making it more effective in perception and decision-making in complex driving environments.
\subsection{Lidar Points}
In the local map perception task of BEV, in addition to using a pure visual surround camera as a single data input, multimodal methods fuse multi-modal information such as lidar point clouds and camera data to perform depth-aware BEV transformation. Compared to single vision and multimodal (RGB+LiDAR) methods, the multimodal fusion method performs excellently in accuracy despite increasing additional computational complexity.
The processing of lidar point cloud data is a crucial step in multimodal perception tasks. The feature extraction of lidar point cloud data in P-MapNet\cite{P-MapNet} requires voxelization of the point cloud first, followed by the use of multi-layer perceptron (MLP) to extract local features of each point. Maximizing pooling selects the largest feature value from multiple local features to form a global feature representation, enhancing the model's global perception ability of point cloud data\cite{Pointnet}.

Given the lidar point cloud P and panoramic images I.
\begin{equation}
    M = F_{2}(F_{1}(P,I)), \label{eq:einstein}
\end{equation}

where $\mathit{F_{1}}$ represents the feature extractor, extracting multimodal inputs to obtain BEV features, $\mathit{F_{2}}$ represents the decoder and outputs the detection results.

The method in MapLite 2.0\cite{Maplite2.0} further integrates lidar point cloud data with data from other sensors and integrates it with coarse road map obtained from SDMap (such as OpenStreetMap\cite{Openstreetmap}) Use the coarse route map information from the SDMap to refine the geometric shape and topological structure of the road. This not only improves the accuracy of the map, but also enhances the understanding of complex road environments\cite{Maplite2.0}. It is also used to generate high-definition maps online by projecting lidar intensity data in bird's-eye view. Integrating multimodal data, not only provides detailed spatial information, but also enables precise semantic segmentation of the driving environment.
\subsection{SDMap}
In the context of enhancing local map perception tasks, incorporating SDMap information as prior knowledge can significantly improve the performance of vision and lidar sensors, particularly in long-distance and occlusion scenarios\cite{SMERF}. To integrate SDMap effectively into network structures while preserving their unique road information, various representations have been explored. SDMap can generally be categorized into two forms: raster and vector.

\begin{figure}[t]
    \centering
    \includegraphics[width=\linewidth]{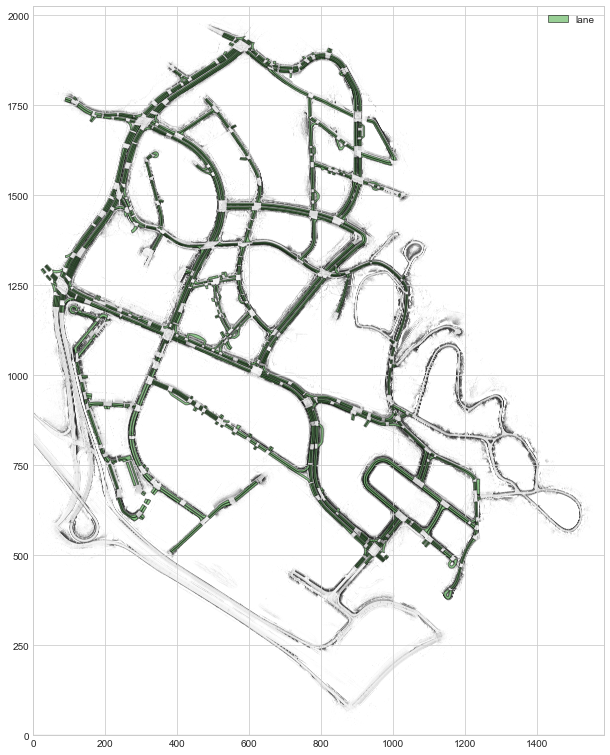}
    \caption{Sample graph of SDMap from nuScenes dataset.}
    \label{fig:sdmap}
\end{figure}

An example of an SDMap is illustrated in Fig. \ref{fig:sdmap}. This figure demonstrates how different forms of SDMap representations can be utilized to supplement the local map construction process, thereby enhancing the overall performance of perception systems.

\begin{equation}
    M = F_{2}(F_{1}(P,L,S)), \label{eq:einstein1}
\end{equation}

Feature extractors can contain multiple modal data. Here S is the SDMap prior that comes in the form of road center-line skeletons. Where $\mathit{F_{1}}$ represents the feature extractor, extracting multimodal inputs to obtain BEV features, $\mathit{F_{2}}$ represents the decoder, and outputs the detection results.


\begin{figure*}[t]
    \centering
    \includegraphics[width=\linewidth]{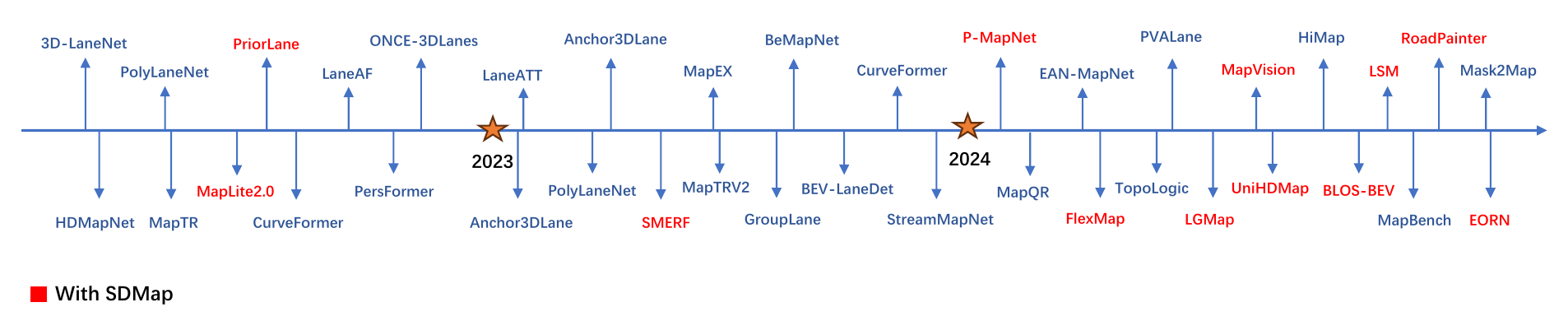}
    \caption{This diagram chronologically presents some of the classic local map construction methods using SDMap in recent years, as well as other methods.}
    \label{fig:time}
\end{figure*}

\subsubsection{Representation of Raster}
MapLite2.0\cite{maplite} was the first to introduce SDMap into local map perception tasks. PriorLane\cite{Priorlane} modeling the map as a binary image, where '1' represents the drivable area and '0' represents the non-drivable area. Similarly, MapVision\cite{MapVision} also uses the one-hot encoding method, then concatenates the position encoding information and extracts the SDMap features through the encoder. The SDMap is aligned with ego data through the KEA module proposed in the algorithm and then fused with sensor data to obtain a mixed expression. Both P-MapNet\cite{P-MapNet} and MapLite2.0\cite{Maplite2.0} use rasterization to represent SDMap, but the difference is that after P-MapNet\cite{P-MapNet}, a CNN is used to extract information from the rasterized SDMap, which is used as a source of additional information for BEV feature optimization (i.e. key and value); MapLite2.0\cite{Maplite2.0} takes SDMap as the initial estimate of HDMap, converts it into a BEV perspective and combines it with images input from sensors. It is trained through a CNN to predict its semantic labels. Finally, these semantic segmentation results are transformed into distance transformations for specific labels, and a structured estimator is used to maintain local map estimates and integrate SDMap priors.

\subsubsection{Representation of Vectors}
SMERF\cite{SMERF} was the first to propose a transformer-based encoder model for road topology inference and introduce a ployline sequence representation and a transformer encoder to obtain the final map representation of the scene. Most subsequent studies followed this approach. Specifically, the roads in SDMap are first abstracted in the form of polylines. For the ployline data, N data points are obtained through uniform sampling. Then, after sin cos encoding, the N-d dimensional line description is obtained. Consider a vertical line with small curvature, which is characterized by very similar x-axis or y-axis values for all points. Directly inputting the coordinates of these points into the model may result in insufficient differentiation of this curvature. 

Therefore, using sine embedding will make this difference more apparent, thereby improving the interpretability of the model for these features. In practice, the coordinates of each line will be normalized to the range of 0 to 2$\pi$ relative to the BEV range before embedding the coordinates of each line. These encoded data will go through several layers of transformers to obtain map feature representations.

\subsubsection{Encoding of Other Information}

In addition to encoding the polyline coordinates of SDMap. SMERF\cite{SMERF} uses one-hot encoding to encode the type of road into a vector with dimension K (the number of road types). For ground elements within the perceptual range, M * (N * d + K) encoded data will be obtained, which will be transformed through several layers to obtain map feature representations. The ablation experiment has shown that adding more road type information can improve the effectiveness of lane detection and road topology inference.
 
\section{Multimodal Fusion Method} \label{sec:mulfusion}
\noindent The method of using only images as input, exemplified by MapTR\cite{MapTR}, based on the encoder-decoder architecture, has established a classic paradigm for local map construction, paving the way for subsequent approaches. MapEX\cite{MapEX} addresses situations where existing map information is incomplete or inaccurate by converting existing map elements into non-learnable queries and combining them with learnable queries for training and prediction. StreamMapNet\cite{Streammapnet} further enhances this by incorporating comprehensive temporal information, significantly improving performance in occluded areas.
3D-LaneNet\cite{3d-lanenet} adopts an end-to-end learning framework, integrating tasks such as image encoding, spatial transformation between image views and top-down views, and 3D curve extraction into a single network. Gen-LaneNet\cite{Gen-lanenet} proposes a two-stage framework that decouples the learning of image segmentation sub-networks and geometric encoding sub-networks. 

Additionally, several monocular 3D lane detection methods, such as\cite{Persformer}, \cite{Anchor3dlane}, and\cite{Bev-lanedet}, focus solely on visual images as input.
Numerous models, including\cite{LaneATT}, \cite{Laneaf}, \cite{BeMapNet}, \cite{Curveformer}, \cite{Grouplane}, \cite{Maptrv2}, \cite{Pivotnet}, \cite{HIMap}, \cite{LaneCPP}, \cite{PVALane}, \cite{img1}, \cite{mapqr}, \cite{MapBench}, \cite{HIMap}, \cite{Mask2Map} and \cite{img2}, also rely solely on visual images. 
On the other hand, HDMapNet\cite{HDmapnet}, a representative multimodal method, integrates point clouds by encoding these features and predicting vectorized map elements in BEV, achieving effective fusion of multi-sensor data. Furthermore, other models, such as\cite{LiLaDet}, \cite{Petrv2}, \cite{Vectormapnet}, \cite{fusion1}, \cite{fusion2}, \cite{fusion4}, and \cite{fusion5}, incorporate point cloud data as additional input. Fig. \ref{fig:time} illustrates the development trends in local map construction over recent years. Considering the cost of constructing high-precision maps, Maplite 2.0\cite{Maplite2.0} was the first to introduce SDMap into local map perception tasks.
 SMERF\cite{SMERF} and P-MapNet\cite{P-MapNet} combine the feature representation of SDMap with camera input features using a multi-head cross-attention mechanism, enabling more effective lane topology inference. To achieve an effective fusion of visual BEV features and SDMap semantic information, BLOS-BEV\cite{BLOS-BEV} has explored various feature fusion methods. Additionally, methods such as PriorLane\cite{Priorlane}, FlexMap\cite{FlexMap}, Bayesian\cite{Bayesian}, TopoLogic\cite{TopoLogic}, LGMap\cite{LGmap}, MapVision\cite{MapVision}, LSM (Leveraging SD Map to Assist the OpenLane Topology)\cite{lsm}, UniHDMap\cite{UniHDMap}, RoadPainter\cite{RoadPainter}, and EORN\cite{Enhancing_Online} integrate SDMap priors to support local map construction, a trend that is gradually gaining traction (Table \ref{tab:tableTab}).

\begin{table*}[t]
\begin{center}
\caption{Performance of using SDmap method on dataset OpenLane-V2} 
\label{tab:tableTab} 
\begin{tabular}{| c | c | c | c | c | c | c | c | c |} 
\hline
Method & SDmap representation & $DET_l$ $\uparrow$& $DET_t$ $\uparrow$& $DET_a$ $\uparrow$& $TOP_{ll}$ $\uparrow$ & $TOP_{lt}$ $\uparrow$& $OLS$ $\uparrow$& $OLUS$ $\uparrow$ \\
\hline
TopoNet+SMERF\cite{SMERF} & vector & 33.4 & 48.6 & - & 7.5 & 23.4 & 39.4 & -  \\
\hline
TopoLogic\cite{TopoLogic} & - & 34.4 & 48.3 & - & 23.4 & 24.4 & 45.1 & - \\
\hline
MapVision\cite{MapVision}  & vector & 39 & 80 & 40 & 38 & 48 & - & 58 \\
\hline
LGmap\cite{LGmap} & vector & 50.74 & - & 55.57 & 46.32 & 53.59 & - & 66  \\
\hline
RoadPainter\cite{RoadPainter} & vector & 36.9 & 47.1 & - & 12.7 & 25.8 &42.6 & - \\
\hline
TopoNet+OSMR\cite{Enhancing_Online} & raster & 30.6 & 44.6 & - & 7.7 & 22.9 & 37.7 & - \\
\hline
TopoNet+OSMG\cite{Enhancing_Online} & vector & 30 & 47.6 & - & 5.4 & 21.3 & 36.7 & - \\
\hline
LSM\cite{lsm} & vector & 49.97 & - & 49.80 & - & - & - & 63.9  \\
\hline
UniHDMap\cite{UniHDMap} & vector & 49.94 & 79.27 & 46.38 & 43.92 & 52.11 & - & 62.81  \\
\hline
\end{tabular}
\end{center}
\end{table*}

\begin{figure}[h]
    \centering
    \includegraphics[width=\linewidth]{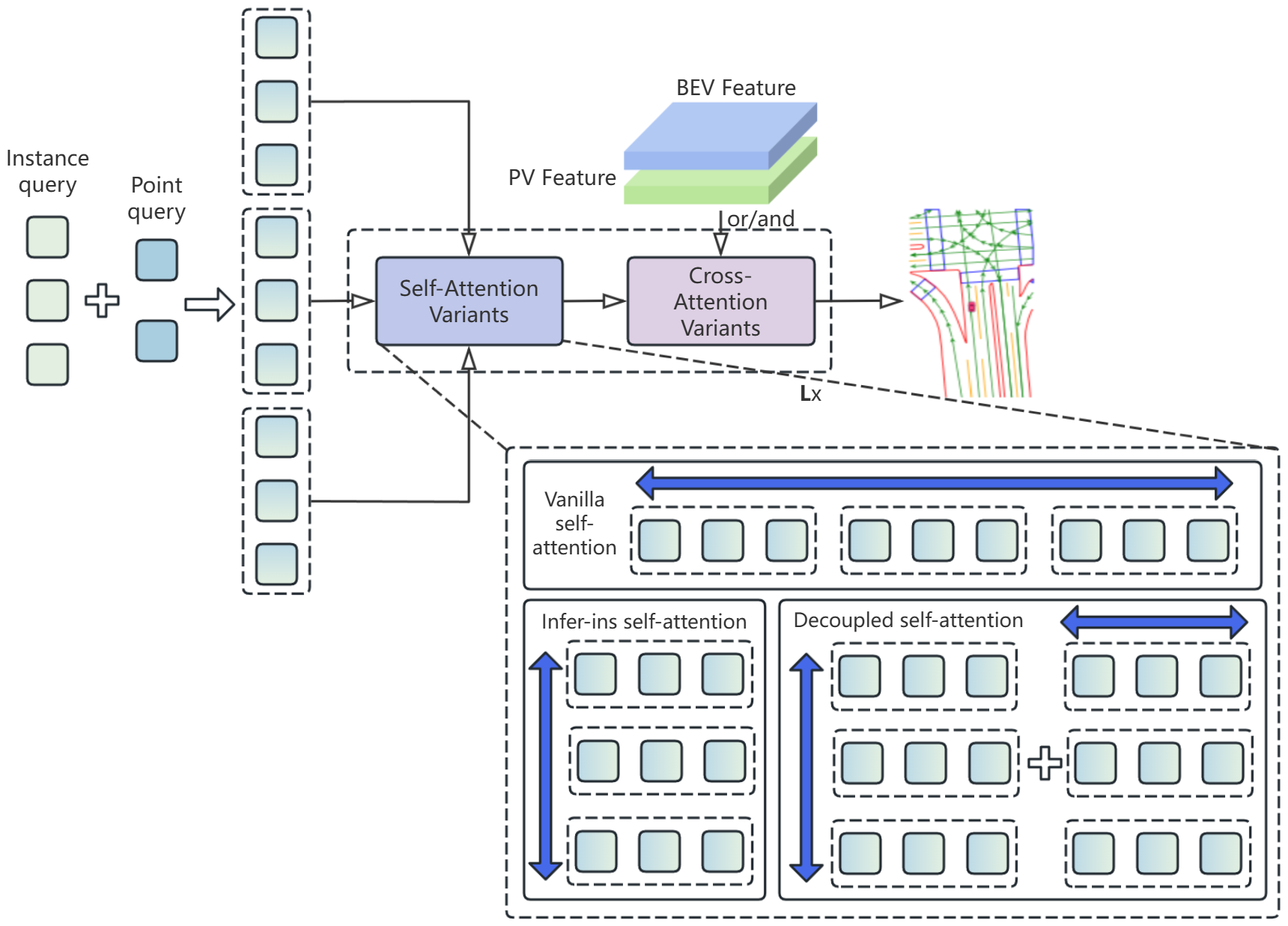}
    \caption{The pipeline of MapTR using self-attention variants and cross-attention variants.}
    \label{fig:MapTR}
\end{figure}

A perspective transition is required before fusion. The focus of this section is to convert feature information extracted from 2D camera sensor images (commonly referred to as PV) into BEV features. Local map perception tasks typically view the ground as a plane, in the Bird's Eye View, Establishing a map in BEV, because on one hand, BEV facilitates information fusion of multiple sensors, and existing advanced BEV object detection work can provide a good foundation. There are both geometric-based and network-based methods for converting the perspective from PV to BEV. Geometric methods can be divided into those based on isomorphic transformations and those based on depth estimation. Network-based methods can be divided into MLP-based and transformer-based methods. The conversion from PV to BEV based on the transformer method can usually be directly achieved using the BEV perception model. MapTR\cite{MapTR} in Fig. \ref{fig:MapTR} uses an optimized GKT\cite{GKT} module based on the View Transformer module in BEVFormer\cite{Bevformer}.

\begin{figure}[h]
    \centering
    \includegraphics[width=\linewidth]{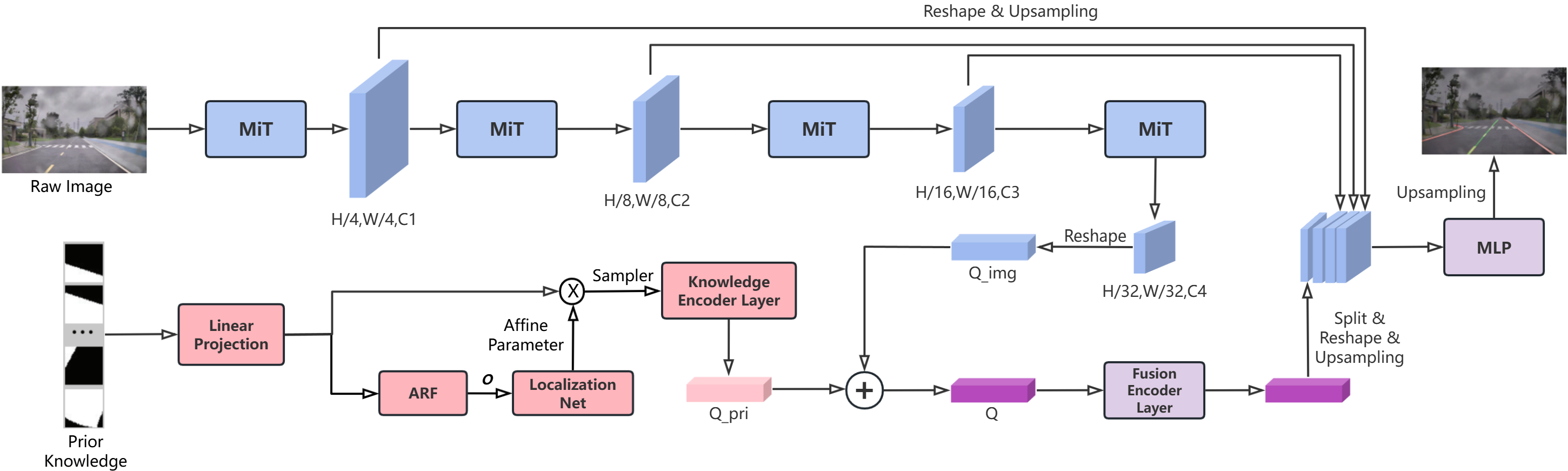}
    \caption{The overview of PriorLane, using prior feature of SDMap to fuse with image feature. }
    \label{fig:PriorLane}
\end{figure}

PriorLane\cite{Priorlane} primarily comprises three sequential components: the MiT block, the KEA module, and the FT block. According to the structure depicted in Fig. \ref{fig:PriorLane}, Unlike typical ViTs\cite{vit}, the MiT block can generate hierarchical features due to its design of overlapping patch merging and efficient self-attention mechanisms. The Knowledge Embedding Alignment (KEA) module is essential for spatially aligning the knowledge embeddings with the coarse vehicle positions in the image features. The fusion transformer is composed of the knowledge encoder layer and the fusion encoder layer working in tandem. The FT block presents an innovative method for detection transformer pre-training, utilizing implicit vision prompts, which in PriorLane\cite{Priorlane} are derived from explicit prior knowledge, with the notable feature that no annotated labels are required for prompt training.

PriorLane\cite{Priorlane} introduces an innovative strategy to enhance lane detection by combining image features with cost-effective local prior knowledge. This prior knowledge is represented as a BEV grid map, segmented into patches, and converted into learnable embeddings through a trainable linear projection. To address the challenge of spatial alignment, PriorLane\cite{Priorlane} incorporates a KEA module, which ensures that the embeddings are accurately aligned spatially, thus preserving consistency between the BEV and camera perspectives. The key innovation of PriorLane\cite{Priorlane} lies in its fusion process, employing a transformer-based architecture to integrate the image features, extracted via an encoder, with the prior knowledge embeddings. Inspired by Segformer\cite{segformer}, PriorLane\cite{Priorlane} utilizes a MLP block to effectively merge the fused features with the original image features, yielding pixel-level segmentation predictions. This methodology has been validated through comparative experiments, demonstrating significant enhancements in lane detection performance by effectively leveraging prior knowledge in conjunction with image features.

MapLite 2.0\cite{Maplite2.0} converts a coarse road graph into an HDmap prior and a rasterized BEV representation. Additionally, the system produces BEV images derived from sensor inputs, which are processed using the CNN trained on existing HDMap data to predict semantic labels. These semantic segmentation results are then converted into distance transforms specific to each label. A structured estimator simultaneously updates the local map and integrates the SDMap prior.
This approach enables the online generation of HDMap, relying solely on SDMap and onboard sensor data. As a result, autonomous vehicles can navigate unmapped areas, while the system also detects environmental changes that could signal the need for map updates.

SDMap provides lightweight priors, requiring minimal storage and bandwidth. While offering less detail than HDMap, SDMap is sufficient for navigation, representing roads as graphs where nodes indicate intersections and edges represent road centerlines, annotated with details like road names, speed limits, and lane counts. In contrast, the HDmap model is layered into road networks, sections with constant lane counts, and directional lanes, ensuring each section connects to specific nodes.

\begin{figure}[h]
    \centering
    \includegraphics[width=\linewidth]{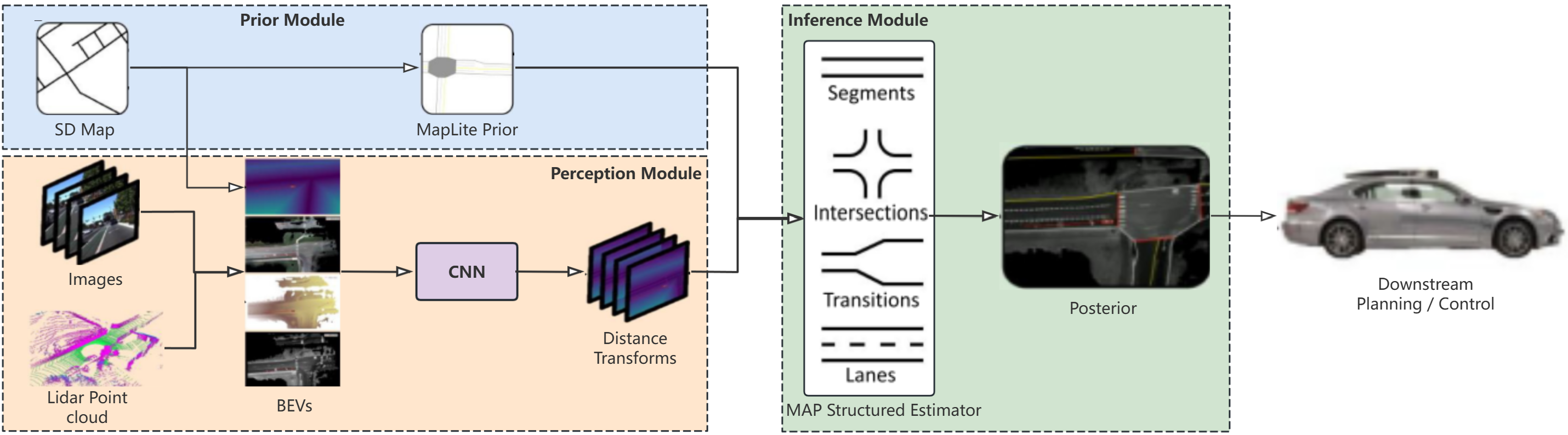}
    \caption{The overview of MapLite 2.0. Using SDMap as prior information, it supports the online inference of local high-definition maps from perception data.}
    \label{fig:MapLite}
\end{figure}

To create BEV features, MapLite\cite{maplite} demonstrates that point clouds from lidar sensors are gathered over time and paired with camera images (Fig. \ref{fig:MapLite}). The points are subsequently projected into a BEV frame by partitioning the space surrounding the vehicle into a grid, which captures intensity information, height, and RGB values for each grid cell. A ResNet-101-based semantic segmentation model is utilized to predict the positions of important map objects such as lane markings and road boundaries. This model is fine-tuned via transfer learning from a DeepLabv3 network\cite{deeplabv3} pre-trained on the COCO dataset\cite{coco}. After segmentation, the output is converted into signed distance transforms, which handle occlusions and remove artifacts caused by missing data.

In the online HDmap estimation process, initial map parameters are modeled as Gaussian distributions. The Prior of MapLite\cite{maplite}, the first HDmap estimate, is constructed offline by comparing SD and HD maps, predicting parameters based on observed distributions. As new sensor measurements arrive, the map state is updated by incorporating the distance transforms from the segmentation step. Optimization is then used to refine parameter estimates, ensuring that sensor data is fully integrated while maintaining an accurate and reliable map estimate.

The SMERF framework\cite{SMERF} enhances an existing lane topology model by integrating priors from SDMap, leading to improved detection of lane centerlines and enhanced relational reasoning. In this context, the SDMap is encoded into a feature representation through a transformer encoder. By applying cross-attention between the feature representation of the SDMap and the onboard camera inputs, the framework constructs BEV features crucial for lane detection and relational reasoning. This pipeline undergoes end-to-end training with the lane topology model, requiring no additional training signals. To leverage the polyline sequence representation of the SDMap, a transformer encoder is employed to develop a feature representation tailored for the downstream lane topology task. The embedding of the polyline sequence occurs through a linear layer, consistent with typical transformer encoder architectures. This process ensures that the discrete one-hot representation of road types is effectively transformed into a continuous space. 
\begin{figure}[h]
    \centering
    \includegraphics[width=\linewidth]{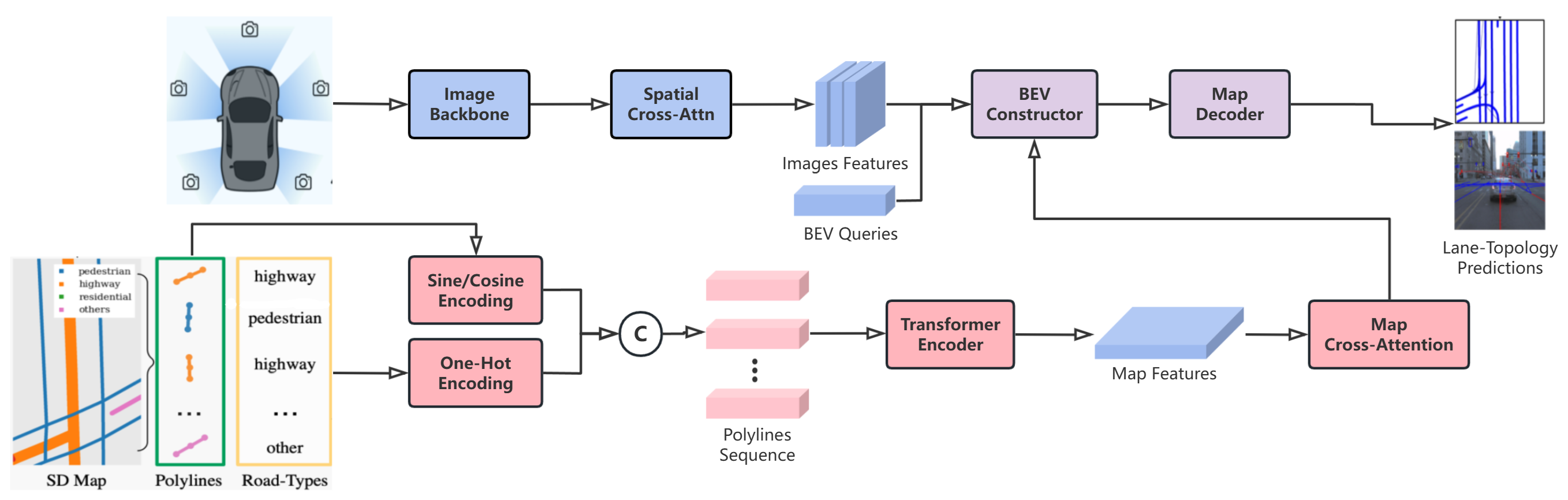}
    \caption{The overview of SMERF. Its SDMap feature is encoded from the polylines and road types of SDMap.}
    \label{fig:SMERF}
\end{figure}

As shown in Fig. \ref{fig:SMERF}, SMERF\cite{SMERF} fuses the features derived from the SDMap with intermediate BEV feature representations by utilizing multi-head cross-attention. This method is adaptable to nearly all transformer-based lane topology models, as it establishes cross-attention between the BEV feature queries and SDMap features at each intermediate encoder layer. The fused BEV features not only incorporate 3D information obtained from images but also capture road-level geometric details extracted from the SDMap. Subsequently, the decoder of the lane topology model takes these SDMap-augmented features as input to predict lane centerlines, traffic elements, and affinity matrices that facilitate the association between lane centerlines and traffic elements. Importantly, the integration of SDMap empowers the model to infer lanes that are distant or even obscured by buildings.

In P-MapNet\cite{P-MapNet}, the localized SDMap data for the corresponding area, obtained from OSM based on onboard GPS information, is transformed into the ego vehicle's coordinate system. However, misalignment issues arise due to the low accuracy of the OSM data and biases in the GPS signals, posing a challenge for the fusion of SDMap priors. Following extraction and rasterization, the rasterized SDMap prior often suffers from spatial misalignment, meaning it does not align precisely with the vehicle's current operational location. This misalignment can be attributed to inaccurate GPS signals or rapid vehicle movement. As a result, the straightforward approach of directly concatenating BEV features with SDMap features becomes ineffective.

\begin{figure}[t]
    \centering
    \setlength{\fboxsep}{0pt}
    \includegraphics[width=\linewidth]{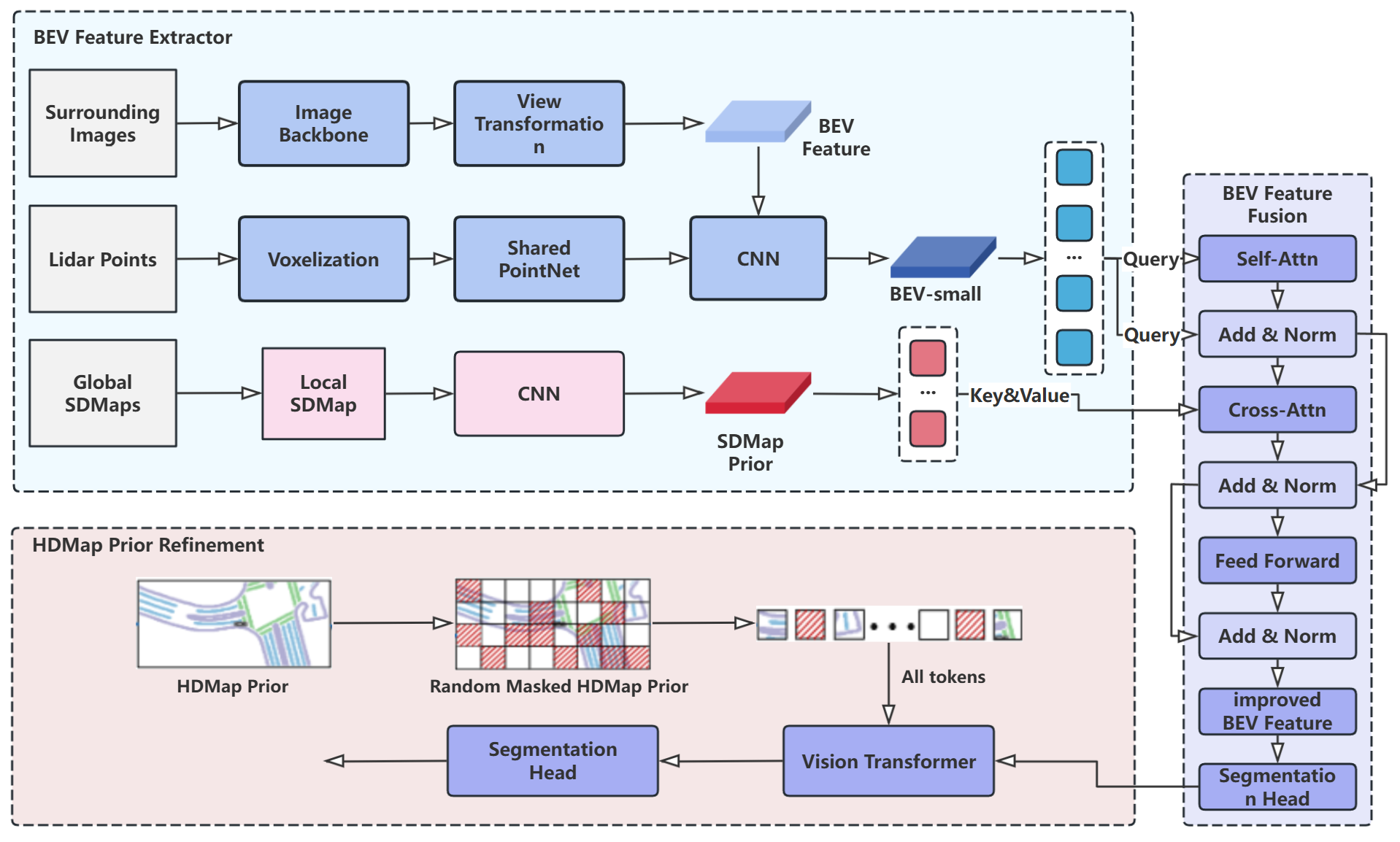}
    \caption{The overview of P-MapNet, including BEV feature extractor module, HDMap prior refinement module and BEV feature fusion module.}
    \label{fig:p-mapnet}
\end{figure}

To address this kind of problem, a multi-head cross-attention module is employed\cite{P-MapNet}, enabling the network to leverage cross-attention to identify the most suitably aligned locations, thus effectively enhancing the BEV features with the SDMap prior in Fig. \ref{fig:p-mapnet}. The SDMap is processed through a convolutional network alongside sine positional embedding, resulting in the generation of SDMap prior tokens. Subsequently, multi-head cross-attention is applied to enrich the BEV queries by incorporating information from the SDMap priors. To improve the continuity and realism of HDMap generation in these scenarios, researchers utilize an adapted pre-trained MAE module\cite{mae} to closely approximate the distribution of the HDMap. This approach effectively captures the inherent distribution, enhancing the overall quality of the generated map.

\begin{figure}[h]
    \centering
    \includegraphics[width=\linewidth]{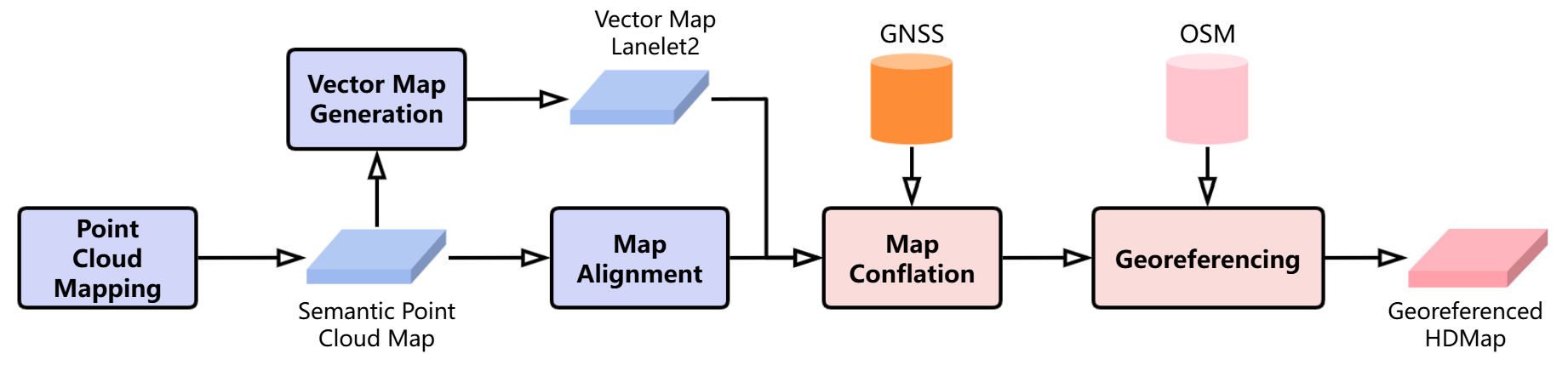}
    \caption{The FlexMap pipeline using GNSS and OSM data.}
    \label{fig:FlexMap}
\end{figure}

FlexMap Fusion\cite{FlexMap} is composed of three distinct modules as illustrated in Fig. \ref{fig:FlexMap}: First, the map alignment module aligns the HDmap to the vehicle’s real-time kinematic (RTK)-corrected GNSS trajectory within a projected, local coordinate frame. Second, the map conflation module merges available data from OpenStreetMap (OSM)\cite{Openstreetmap} into a vector map (VM). Lastly, in the georeferencing module, local coordinates are projected into global coordinates to produce a georeferenced VM. The following sections detail the implementation of each FlexMap Fusion module.

In the Map Alignment module, the HDmap is synchronized with the vehicle’s RTK-corrected GNSS trajectory in a projected local coordinate frame. To minimize any shift between these trajectories, the initial GNSS measurement is used as the projection origin. A rigid transformation is employed to account for any remaining rotation or translation discrepancies. To correct residual geometric deviations, typically caused by accumulated errors during SLAM, a piecewise linear rubber-sheet transformation is applied.

Once the Map Alignment module has been completed, the HD and OSM maps are mostly geometrically aligned, with only minor deviations. A preprocessing step is applied to the lane map for refinement. A road network matching algorithm is then used to identify corresponding objects between the vector map (VM) and OSM. After identifying these correspondences, relevant information from OSM is transferred to the lane map. During the map alignment process, the SLAM trajectory and PCM / VM are aligned with the projected GNSS trajectory. To enable georeferencing, the UTM projection applied during Map Alignment is reversed, converting the local coordinates into global ones.

\begin{figure}[h]
    \centering
    \includegraphics[width=\linewidth]{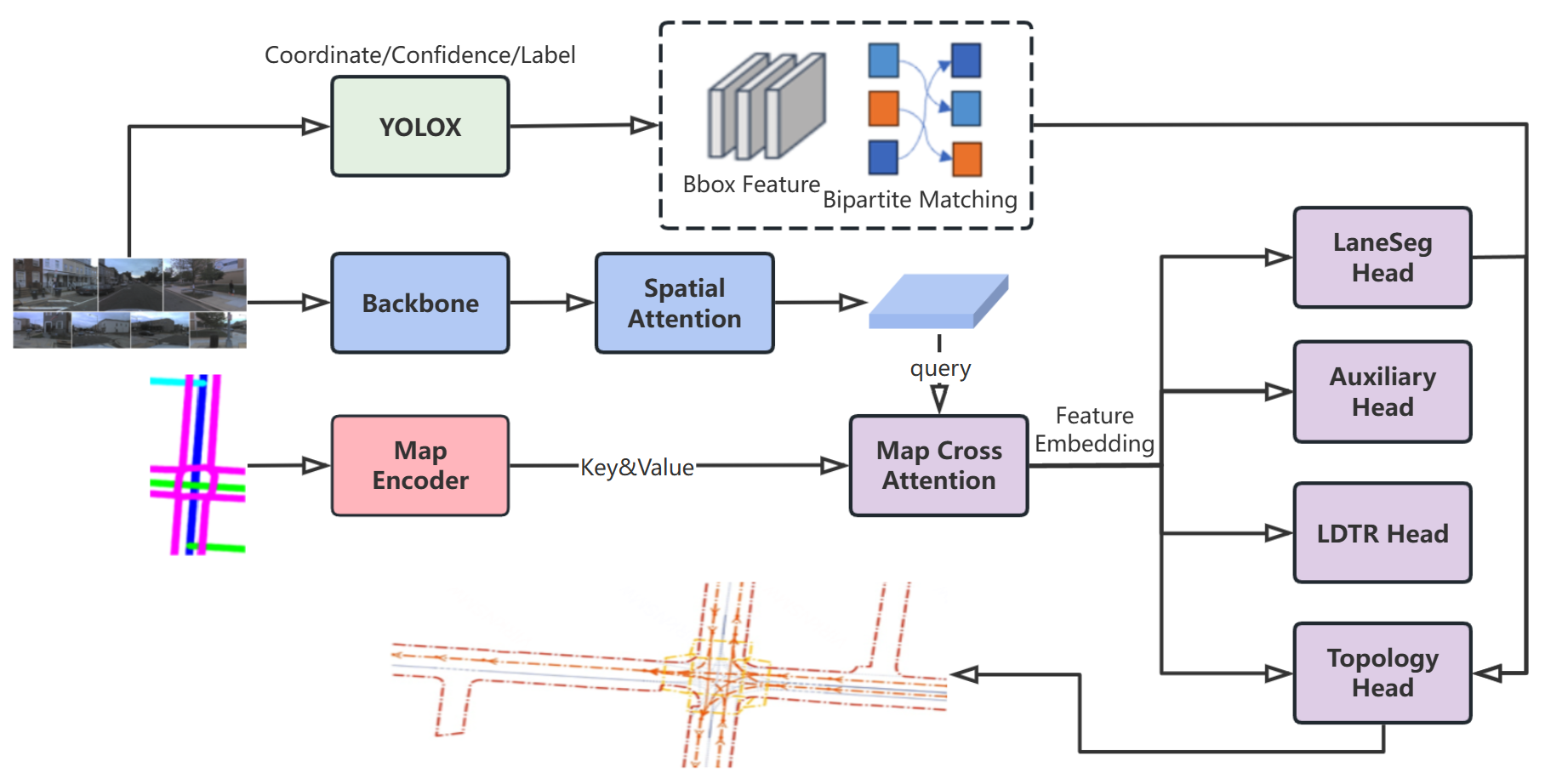}
    \caption{The overview of MapVision, which uses multi-heads.}
    \label{fig:mapvision}
\end{figure}

Many current algorithms build BEV features from multi-perspective images and utilize multi-task heads to identify road centerlines, boundary lines, and other areas. However, performance often diminishes at the far ends of roads and faces difficulties when the main subject in the image is obscured. To tackle these issues, It can be seen from Fig. \ref{fig:mapvision} that MapVision\cite{MapVision} incorporates both multi-perspective images and SDMap as inputs.
SDMap is utilized as a supplementary element, enhancing the understanding of road topology from a BEV perspective while providing map priors over extended distances.

The framework of MapVision\cite{MapVision} retains the structure of SMERF\cite{SMERF}. In this setup, lane lines are encoded using sine-cosine position encoding, while category information is represented through one-hot encoding. These features are then concatenated and processed by a transformer encoder to extract meaningful map features.
Informed by BEVFormer\cite{Bevformer}, PV features are projected onto BEV, followed by the fusion of image features with SDMap features within the BEV domain. Moreover, auxiliary foreground segmentation tasks are introduced for both PV and BEV to improve feature extraction capabilities.
For the cross-attention mechanism, SDMap features are employed as the key and value, while the feature map derived from multi-perspective images serves as the query. The integration of SDMap priors results in a marked improvement in the quality of HDmap construction.
To further enhance the map encoder's comprehension of the map, a pre-training phase is conducted. This pre-training enables the map encoder to learn the conversion process from the position-encoded SDMap to the feature map output, significantly advancing topological reasoning capabilities.

LGMap\cite{LGmap} introduces a symmetric view transformation that combines forward and backward projections to leverage their complementary advantages. The Lift-Splat-Shoot\cite{lss} approach capitalizes on depth distribution to model the uncertainty of each pixel’s depth and dense BEV features from BEVFormer\cite{Bevformer}, using depth supervision solely during the training phase. This helps mitigate false correlations between 3D and 2D spaces caused by occlusions. By employing sinusoidal embedding, BEVFormer\cite{Bevformer} applies cross-attention between the SDMap feature representation and the features from visual inputs at each encoder layer.
\begin{figure}[h]
    \centering
    \includegraphics[width=\linewidth]{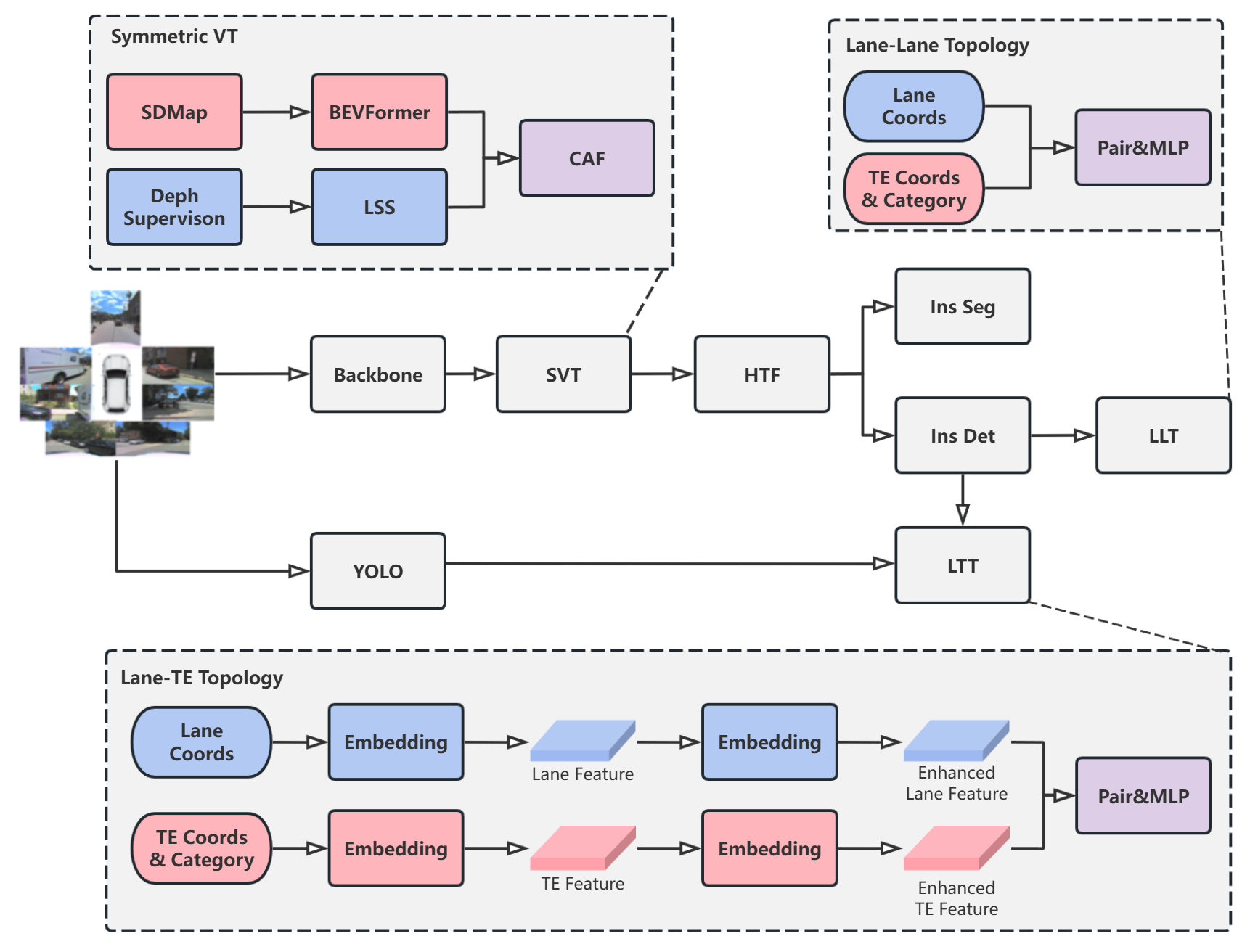}
    \caption{The pipeline of LGmap, including symmetric VT module, Lane-TE topology module and Lane-Lane topology module.}
    \label{fig:LGmap}
\end{figure}

In Fig. \ref{fig:LGmap}, Hierarchical Temporal Fusion (HTF) fully utilizes the local fusion capabilities of a streaming strategy while also harnessing the long-range fusion strengths of a stacking strategy. This method effectively minimizes memory usage and latency costs compared to the stacking approach. To accommodate various map elements with distinct shape priors, the instance-wise detection decoder is extended with additional segmentation tasks.

Extending the range of BEV representation offers substantial benefits for downstream tasks, including topology reasoning, scene understanding, and planning. This extension provides more comprehensive information and enhances reaction times. The SDMap serves as a lightweight representation of road structure topology, notable for its ease of acquisition and low maintenance costs. 

By merging close-range visual data from onboard cameras with beyond line-of-sight (BLOS) environmental priors from SDMap, perceptual capabilities can be expanded to a range of 200 meters.
For the BEV feature extraction baseline, LSS\cite{lss} is chosen for its lightweight and efficient design, which facilitates easy integration. The backbone of the SDMap Encoder is built upon a VGG architecture.

\begin{figure}[h]
    \centering
    \includegraphics[width=\linewidth]{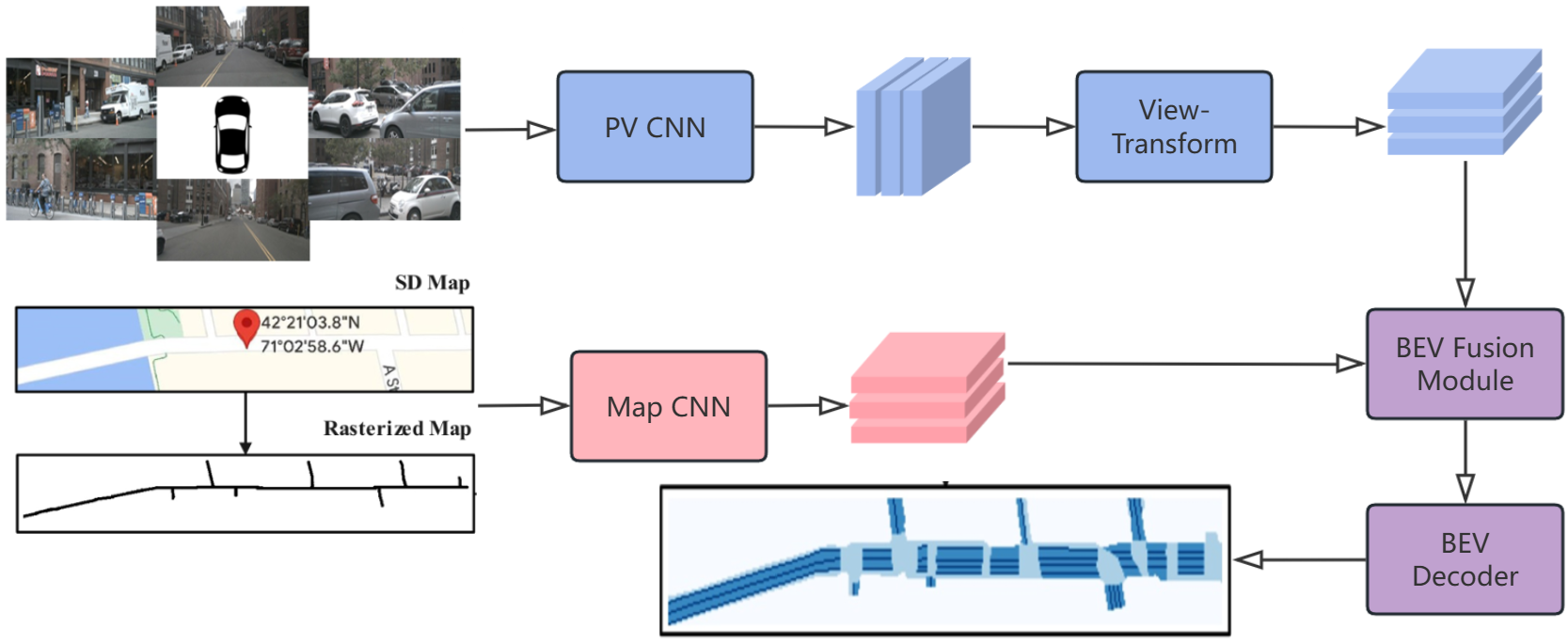}
    \caption{The pipeline of BLOS-BEV, where the rasterized map is used to generate SDMap features by CNN to participate in feature fusion. }
    \label{BLOS}
\end{figure}

As illustrated in Fig. \ref{BLOS}, BLOS-BEV\cite{BLOS-BEV} is currently investigating various fusion strategies to combine visual BEV features with SDMap semantics, aiming for optimal representation and performance. The three main approaches being explored are addition, concatenation, and the cross-attention mechanism.
Within the BEV Decoder, high and low-resolution fused features are received. The low-resolution features undergo upsampling by a factor of 4 to ensure alignment in height and width with the high-resolution features. Following this, the two sets of features are concatenated along the channel dimension. This is then processed through two convolutional layers and additional upsampling to produce the final BEV segmentation map.

UniHDMap\cite{UniHDMap} introduces a unified detection framework that integrates information from lanes, pedestrian crossings, and road boundaries. It modifies LaneSegNet\cite{Lanesegnet} for road segmentation, ensuring a unified representation of all road detection elements. To accelerate convergence and improve detection accuracy, it adopts a one-to-many approach by increasing the number of queries fivefold for additional supervision. YOLOv8\cite{YOLOv8} is utilized for 2D traffic detection, while two MLP-based heads are employed for lane-lane and lane-traffic topology prediction, enhancing the model's ability to manage complex road scenarios.

\begin{figure}[h]
    \centering
    \includegraphics[width=\linewidth]{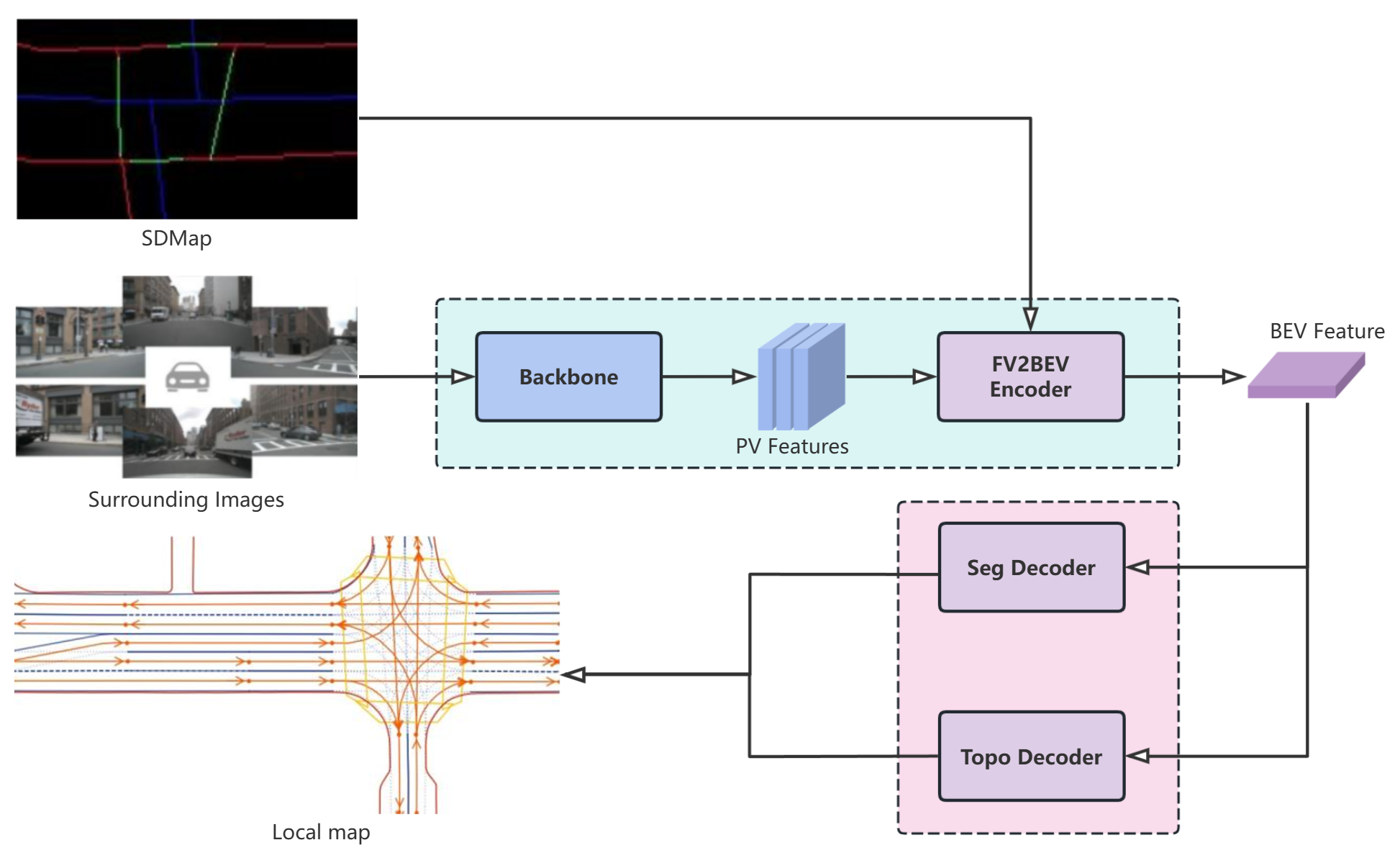}
    \caption{The pipeline of UniHDMap.}
    \label{fig:UniHDMap}
\end{figure}

In Fig. \ref{fig:UniHDMap}, UniHDMap\cite{UniHDMap} utilizes a ResNet backbone\cite{ResNet} to extract feature maps from images, followed by a view conversion through BEVFormer's PV-to-BEV encoder module. The transformer-based detection decoders collect BEV features and update queries across layers. During this process, vector-encoded information from standard-definition maps is injected to provide location guidance for map elements. SDMap is converted into vector polyline sequences, which interact with BEV queries via cross-attention after being encoded by a transformer encoder, producing enhanced BEV features for improved map-based detection.

\begin{figure}[h]
    \centering
    \includegraphics[width=\linewidth]{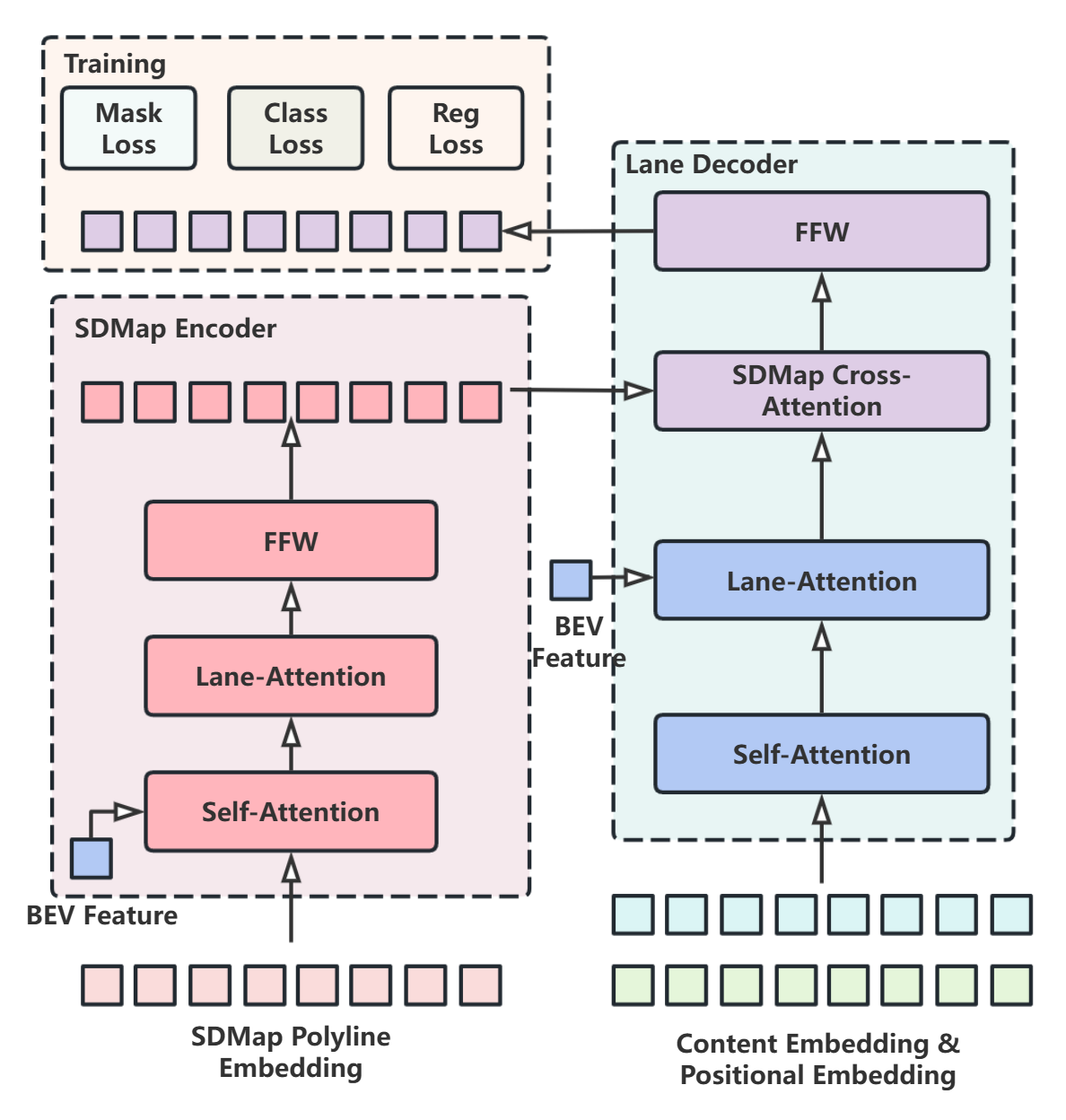}
    \caption{The pipeline of LSM, including SDMap Encoder and Lane Decoder.}
    \label{fig:LSM}
\end{figure}

LSM\cite{lsm} constructs BEV features from multi-view input images using BEVFormer\cite{Bevformer}. An SDMap encoder then extracts features from the SDMap elements, which are represented as polylines. As shown in Fig. \ref{fig:LSM}, polylines from the SDMap are encoded into embeddings, followed by cross-attention to fuse the SDMap features with the BEV features. 

Additionally, a novel ensemble method is introduced, which decouples topological relationships from road element detection using an MLP. This improves the accuracy of topology prediction, further enhancing model performance in lane detection tasks.

\begin{figure}[h]
    \centering
    \includegraphics[width=\linewidth]{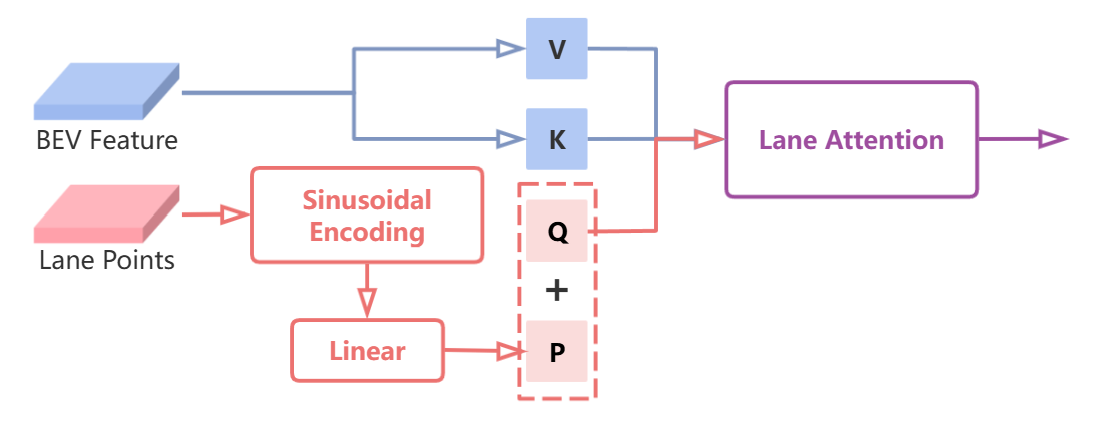}
    \caption{The dynamic positional encoding scheme of LSM.}
    \label{fig:xiaomi_dynamic_pe}
\end{figure}

A dynamic positional encoding scheme is proposed for each lane attention layer in the lane decoder, as shown in Fig. \ref{fig:xiaomi_dynamic_pe}. This scheme updates the positional encoding by utilizing the location information of lane points derived from the output of the preceding decoder layer. This dynamic update ensures that the positional encoding adapts based on the current state of the lane points, improving the model’s ability to capture and predict the topological structure and spatial relationships of lane elements during decoding.

To enhance performance, an ensemble strategy is applied, starting with the results from the best-performing detection model on the validation set as the base proposals. Other candidate models' outputs are integrated using a trust-based voting strategy. Depending on the similarity between candidate proposals and base ones, confidence scores are assigned. This approach reduces false negatives and strengthens true positive detection confidence.

RoadPainter\cite{RoadPainter} processes surrounding images and optionally incorporates SDMap to generate BEV features. These features are used to detect lane centerlines as geometric points in the BEV space and determine their associations, improving the accuracy and reliability of lane topology. RoadPainter\cite{RoadPainter} enhances this process by utilizing an SDMap interaction module, which augments BEV features with road shape priors and beyond-visual-range data extracted from the SDMap. The SDMap is vectorized and converted into BEV features by filling each grid cell with the corresponding semantic type embedding if occupied.

For extracting lane centerlines, RoadPainter\cite{RoadPainter} employs a transformer decoder with a hybrid attention layer, which integrates masked cross-attention, deformable cross-attention, and self-attention. 
To achieve accurate topological associations, it additionally employs a topology head that utilizes centerline instance queries and positional embeddings.

\begin{figure}[h]
    \centering
    \includegraphics[width=\linewidth]{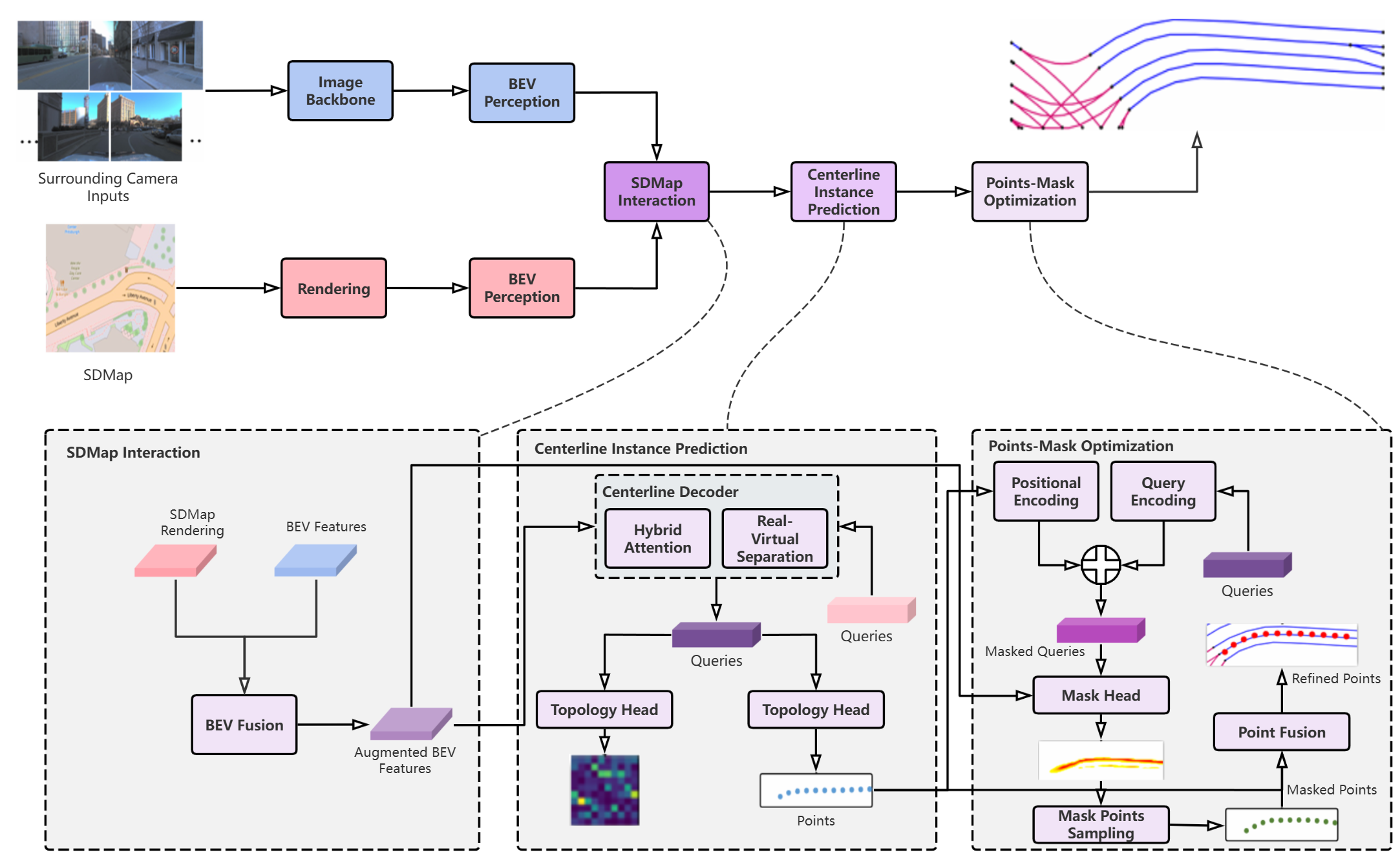}
    \caption{The pipeline of RoadPainter, which shows three important modules.}
    \label{fig:RoadPainter}
\end{figure}

To refine the detected centerline points, RoadPainter\cite{RoadPainter} introduces a points-mask optimization module as illustrated in Fig. \ref{fig:RoadPainter}, consisting of two submodules: points-guided mask generation and points-mask fusion. By leveraging detected centerline points to guide the generation of mask queries, this method addresses accuracy issues, particularly in areas of high curvature. The first stage, mask points sampling, selects a set of points from the generated mask. In the second stage, points fusion combines these sampled mask points with the detected centerline points to produce refined centerline positions. This refinement approach ensures more precise and reliable lane topology reasoning.

\begin{figure}[h]
    \centering
    \includegraphics[width=\linewidth]{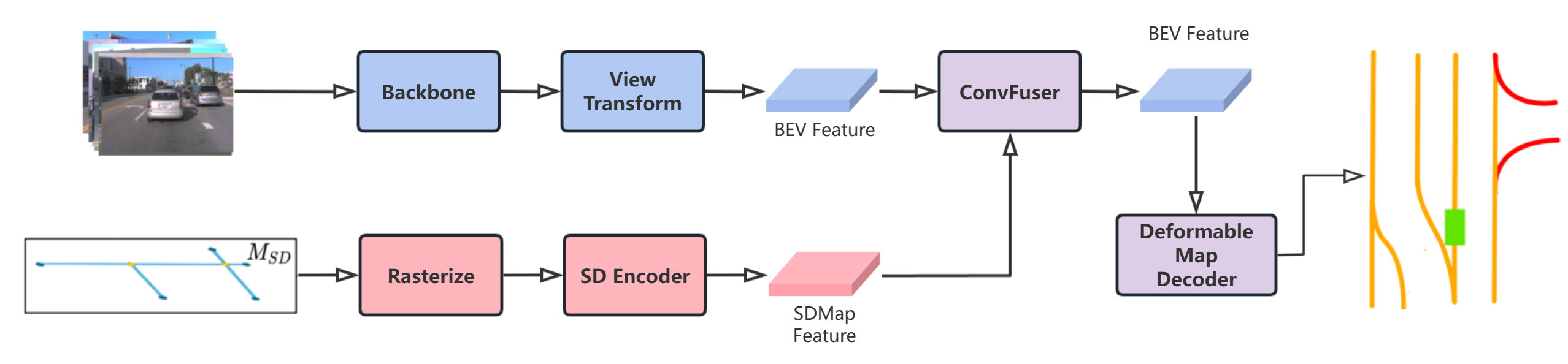}
    \caption{The overview of ENOR.}
    \label{fig:ENOR}
\end{figure}

EORN\cite{Enhancing_Online} centers on utilizing lightweight and scalable SDMap to develop online vectorized HDmap representations. Initially, the integration of rasterized SDMap into various online mapping architectures is explored. A key takeaway from this process is that SDMap encoders are model-agnostic, allowing for rapid adaptation to new architectures that use BEV encoders. Fig. \ref{fig:ENOR} demonstrates that incorporating SDMap as priors in online mapping tasks significantly accelerates convergence and improves the performance of tasks like online centerline perception.

To evaluate the utility of SDMap as priors in HD mapping, EORN\cite{Enhancing_Online} integrates them into recent online mapping efforts. These tasks fall into two categories: the first focuses on perception, which involves detecting map elements such as lane lines, road boundaries, crosswalks, and centerlines. The second area expands perception to include reasoning, detecting traffic elements like lights and signs, as well as understanding their relationships.
In terms of architecture, EORN\cite{Enhancing_Online} incorporates SDMap into SOTA HD mapping systems based on MapTR\cite{MapTR}. An image encoder transforms surround-view images into PV features, which are further processed into unified BEV features using a BEV view transform module. A ResNet-18\cite{ResNet} is employed as the SD encoder to extract SDMap features, which are then interpolated and concatenated with the BEV features from the images along the channel dimension, ensuring proper spatial alignment and boosting the mapping performance.

\subsection{Alignment Methods}
Due to the inherent errors in GPS signals and the influence of vehicle movement, both vectorized and rasterized SDMap priors inevitably have spatial misalignment with the current BEV space, making it difficult to fully align the two. Therefore, before fusion, it is necessary to spatially align the SDMap prior with the current BEV operating space. FlexMap\cite{FlexMap} uses SLAM trajectory and corrected RTK trajectory to calculate the offset and achieve spatial alignment.

\begin{figure}[h]
    \centering
    \includegraphics[width=\linewidth]{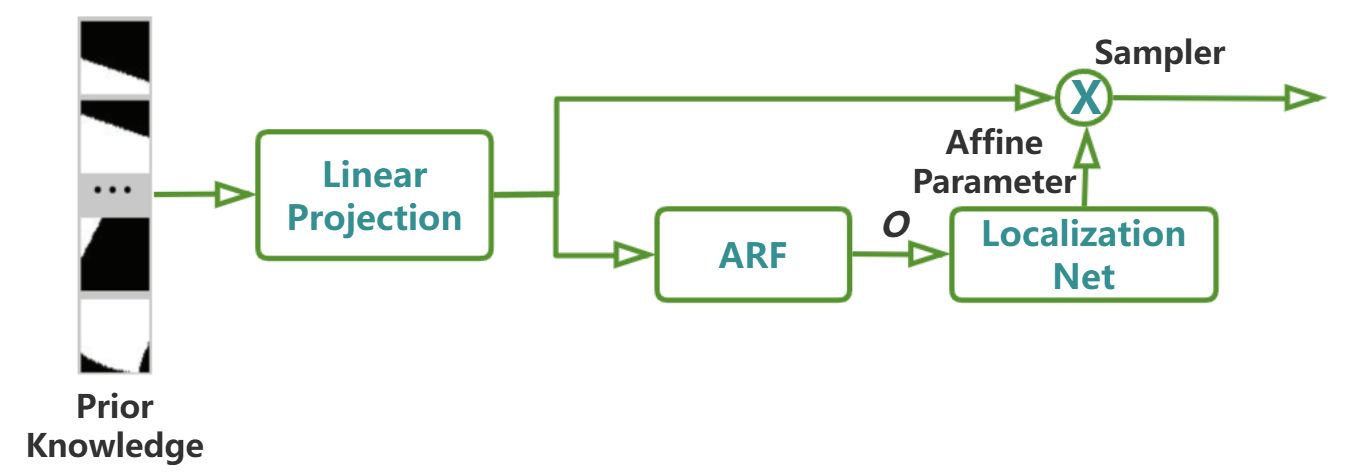}
    \caption{The alignment method of PriorLane.}
    \label{fig:PriorLane_align}
\end{figure}

To solve this problem, PriorMap\cite{Priorlane} sets up a KEA module to embed SDMap prior knowledge and align it with image features in space shown in Fig. \ref{fig:PriorLane_align}. Specifically, first, a feature extraction network is used to extract feature points from the image and feature points from the SDMap prior knowledge. 
Subsequently, these feature points are spatially matched using an alignment algorithm based on attention mechanisms. 

\begin{figure}[h]
    \centering
    \includegraphics[width=\linewidth]{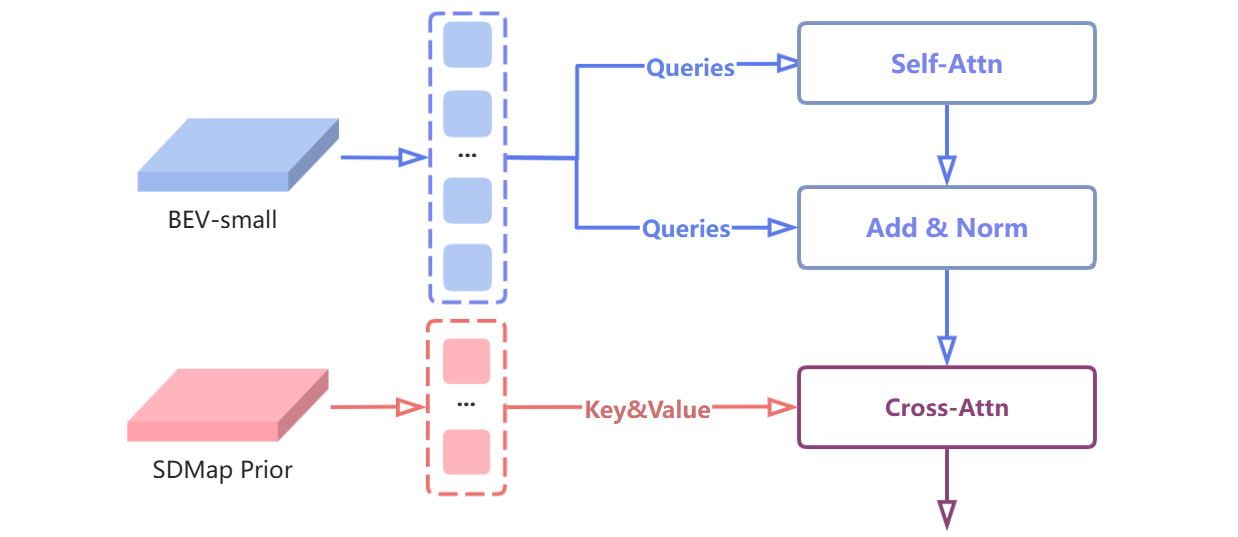}
    \caption{The alignment and fusion methods of P-MapNet using cross attention.}
    \label{fig:mapnet_align}
\end{figure}

Finally, the aligned feature points are further processed through a fusion transformer network and enhance the accuracy and robustness of local map perception algorithms. Similarly, P-MapNet\cite{P-MapNet} first downsamples the rasterized SDMap prior and then introduces a multi-head cross-attention module that allows the network to use cross-attention to determine the most suitable alignment position, effectively enhancing BEV features using SDMap prior, As illustrated in Fig. \ref{fig:mapnet_align}. 

P-MapNet's ablation experiment\cite{P-MapNet} shows that even in the case of weak alignment with BEV space, directly concatenating SDMap prior information still improves the performance of the model. On this basis, adding CNN modules and multi-head cross-attention modules can further improve the performance of the model. This demonstrates the important role of SDMap prior information in local map perception tasks, even without strict alignment, simply adding rasterized SDMap priors can improve model performance. 

In LSM\cite{lsm}, to achieve cross-modality alignment, an additional SDMap cross-attention layer is introduced after each lane attention layer. This layer incorporates SDMap features, enhancing the model's ability to align the spatial context between the SDMap and BEV features, leading to more accurate lane detection and topological reasoning.

\subsection{Fusion Methods}

After obtaining multi-sensor data feature representations, fusion processing is required to obtain stronger feature representations.

\begin{figure}[h]
    \centering
    \includegraphics[width=\linewidth]{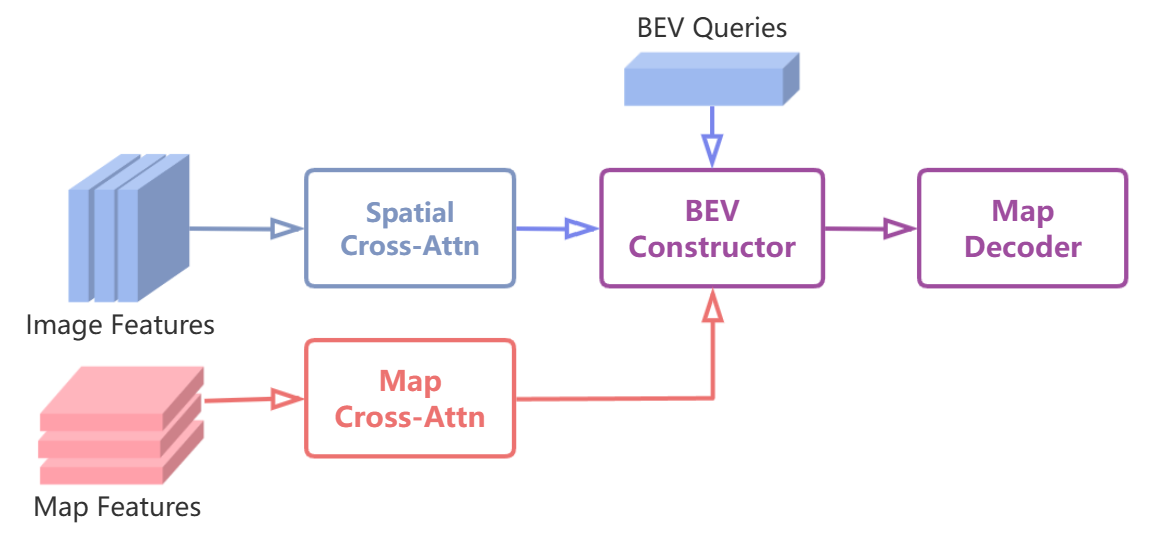}
    \caption{The fusion method of SMERF.}
    \label{fig:SMERF_fusion}
\end{figure}

To align the features of different sensors, it is necessary to achieve fusion in BEV-level features. The image BEV features are obtained from the rounded image through the perspective conversion module. In SMERF\cite{SMERF}, the SDMap feature interacts with the BEV feature through a cross-attention mechanism as demonstrated in Fig. \ref{fig:SMERF_fusion}. 

Firstly, the BEV feature is encoded into a query vector and initialized through a self-attention mechanism. Given the SDMap of the scene, LGMap\cite{LGmap} evenly samples along each of the polylines for a fixed number of points. With sinusoidal embedding, BEVFormer\cite{Bevformer} applies cross-attention between the SDMap feature representation with features from vision inputs on each encoder layer. The SDMap features are encoded into key and value vectors, which are then computed through cross-attention to obtain the final fused camera, BEV features of SDMap.

\begin{figure}[h]
    \centering
    \includegraphics[width=\linewidth]{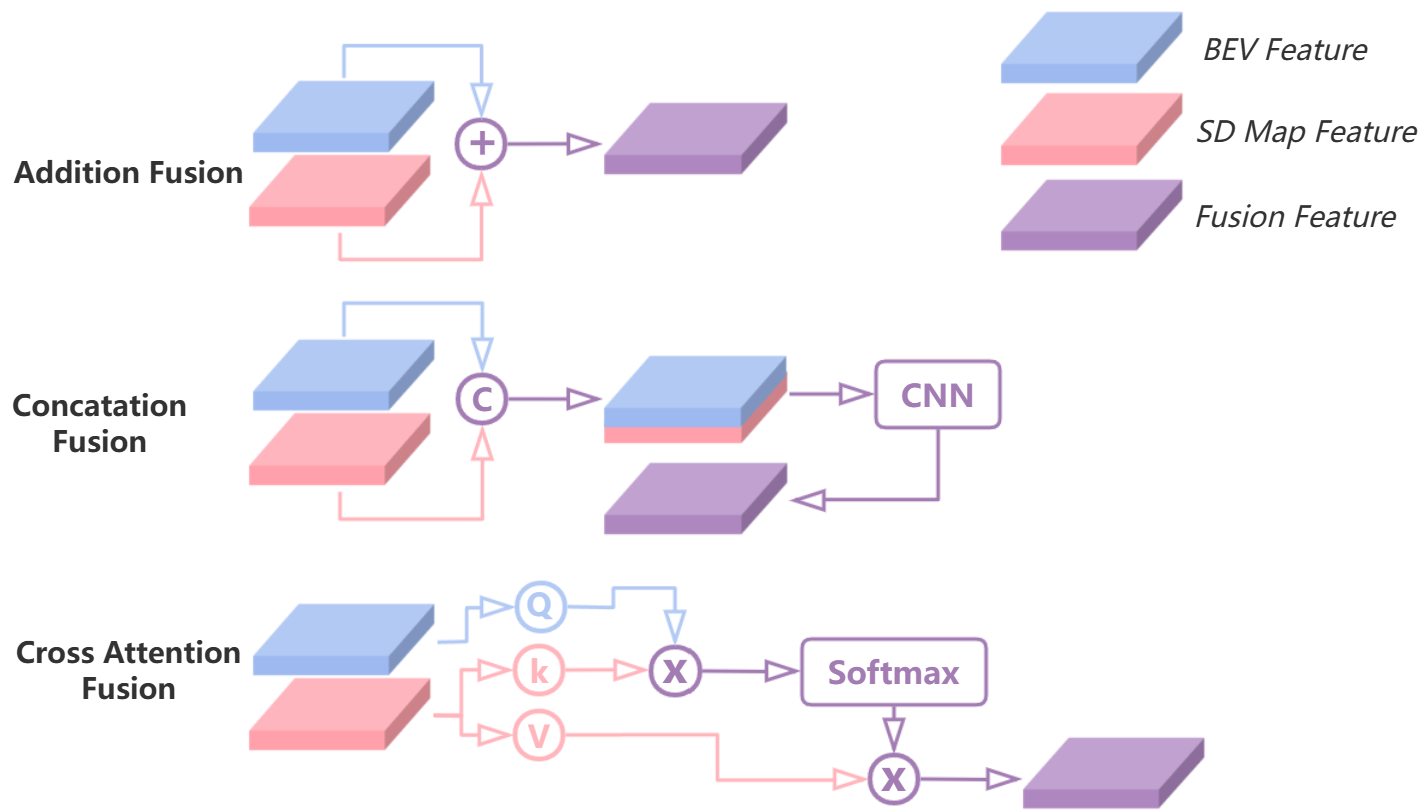}
    \caption{The three kinds of fusion methods in BLOS-BEV.}
    \label{fig:blos_fusion_method}
\end{figure}

In addition to the common fusion method of attention mechanisms, BLOS-BEV\cite{BLOS-BEV} shows in Fig. \ref{fig:blos_fusion_method}, explore different fusion schemes that combine visual BEV features and SDMap semantics to achieve optimal representation and performance, three SDMap fusion techniques were explored: addition, concatenation, and cross-attention. Although all fusion methods outperform those that do not use SDMap, the cross-attention fusion of SDMap performs the best on the nuScenes\cite{nuscenes} and Argoverse datasets\cite{Argoverse}, demonstrating excellent generalization performance and outstanding performance over long distances (150-200m).

In P-MapNet\cite{P-MapNet}, point cloud information has been added, and the lidar point cloud has been voxelated and MLP processed to obtain the feature representation of each point, resulting in lidar BEV. Fusion of Image BEV and lidar BEV are used to obtain further fused BEV features. Further convolutional downsampling of the fused BEV features can alleviate the misalignment problem between image BEV features and lidar BEV features.

\begin{figure}[h]
    \centering
    \includegraphics[width=\linewidth]{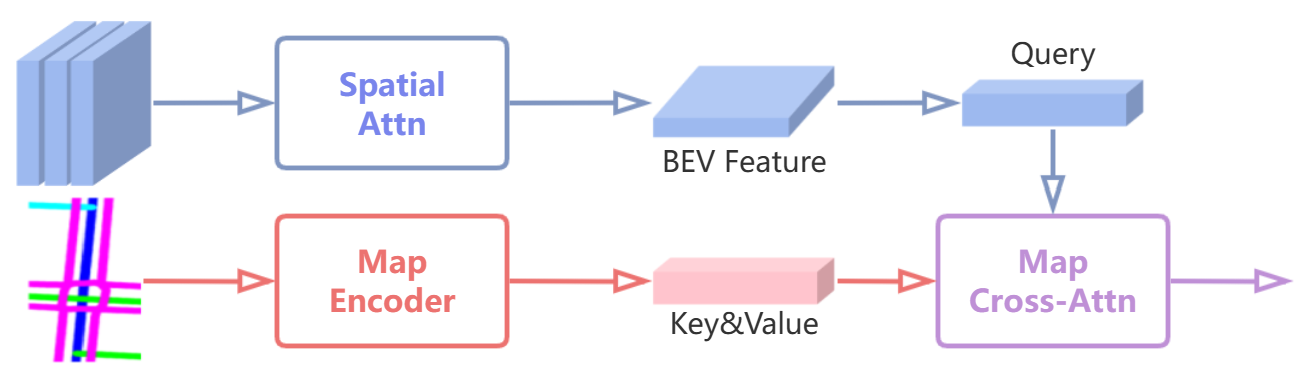}
    \caption{The fusion method of MapVision using attention modules.}
    \label{fig:mapvision_fusion}
\end{figure}

Then, through cross attention mechanism, the SDMap feature interacts with the fused BEV feature, resulting in the final fusion of images and point cloud, BEV features of SDMap. Similarly, MapVision\cite{MapVision} uses the SDMap feature as the key and value, and the feature map formed from multi-perspective images as the query to perform cross-attention (Fig. \ref{fig:mapvision_fusion}).

Similar to LSM\cite{lsm}, a dedicated SDMap encoder is developed to encode vectorized map elements, enriched by BEV features to provide enhanced spatial context. This encoding helps incorporate crucial road layout information and facilitates a more precise understanding of lane detection and traffic elements in the scene.

\begin{figure}[h]
    \centering
    \includegraphics[width=\linewidth]{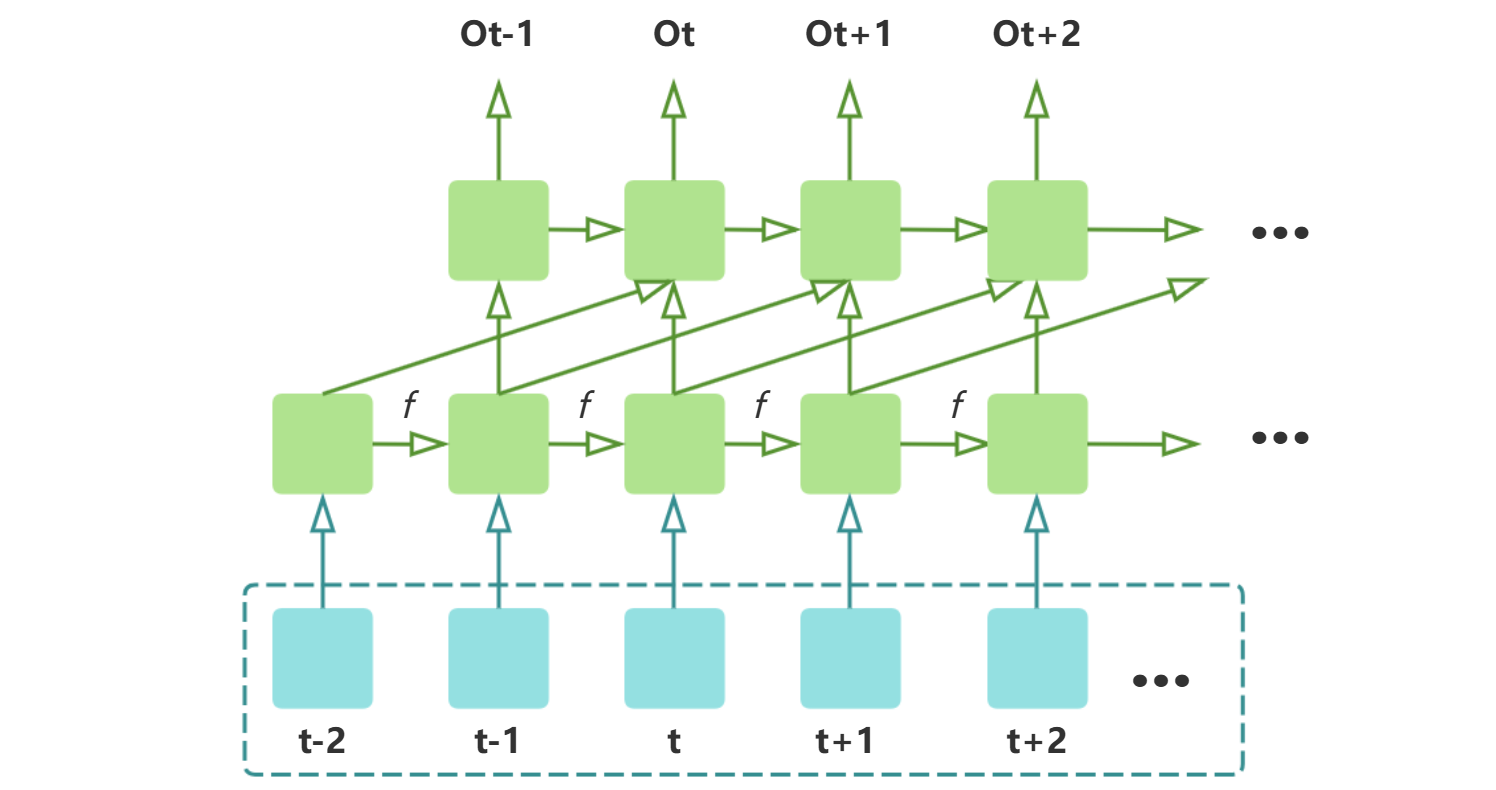}
    \caption{The fusion method of LGmap: Streaming-stacking strategy.}
    \label{fig:LGmap_fusion}
\end{figure}

Fig. \ref{fig:LGmap_fusion} shows the fusion method of LGmap\cite{LGmap}.
The stacking strategy and streaming strategy are the same as StreamMapNet’s summary\cite{Streammapnet}. To demonstrate the effectiveness of long-range stacking for the streaming-stacking strategy in the figure, the stacking previous frame interval parameter is set to 2. The stacking strategy only fuses one previous frame in this figure. It may fuse more than one frame.

\begin{figure}[h]
    \centering
    \includegraphics[width=\linewidth]{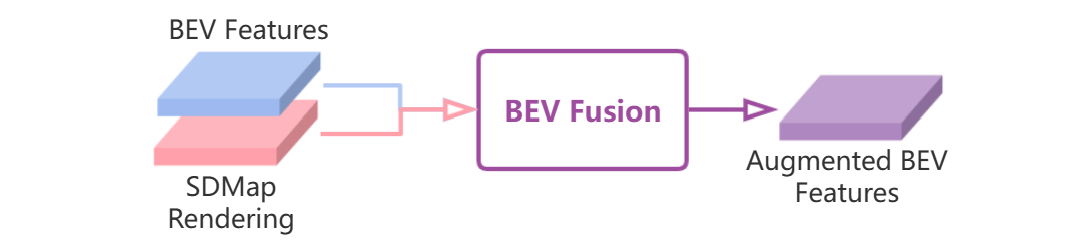}
    \caption{The fusion method of RoadPainter using BEV fusion module.}
    \label{fig:roadpatiner_fusion}
\end{figure}

To address these issues such as occlusion and limited sensing range can result in inaccuracies, RoadPainter\cite{RoadPainter} presents a novel SDMap interaction module that effectively augments BEV features by incorporating beyond-visual-range in Fig. \ref{fig:roadpatiner_fusion}.

\begin{figure}[h]
    \centering
    \includegraphics[width=\linewidth]{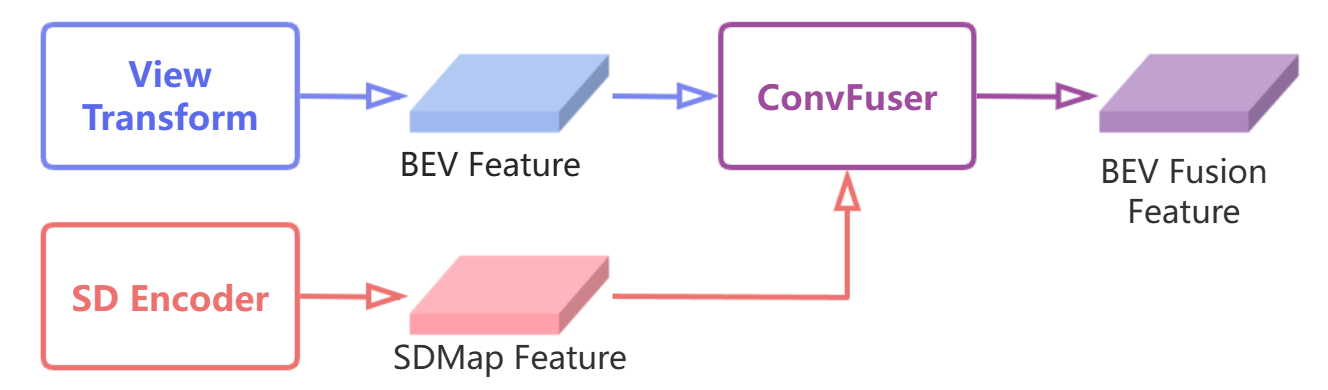}
    \caption{The fusion method of EORN using Convfuser module.}
    \label{fig:eorn_fusion}
\end{figure}

EORN\cite{Enhancing_Online} rasterizes and generates SDMap in BEV. An SD encoder based on a ResNet-18\cite{ResNet} to extract SDMap features. The SDMap feature is then interpolated and concatenated with the BEV feature from images BEV along the channel dimension. fusion method uses a simple two-layer convolutional neural network, the ConvFuser, which fuses concatenated features and outputs the fused BEV features. Another method involves a graph-based encoder that fuses SDMap graphs with BEV features and combines these with the outputs from the centerline deformable decoder using a multi-head attention mechanism.

Fig. \ref{fig:eorn_fusion} shows the fusion method of EORN\cite{Enhancing_Online}.
The subsequent decoder can calculate and output corresponding results for different tasks through query queries from BEV features containing rich information.

\section{Challenges and Future Prospects} \label{sec:challenges}
\noindent Despite significant progress in local map construction with SDMap, several challenges remain that hinder the full realization of its potential. Addressing these issues will require further innovation, interdisciplinary collaboration, and the development of new methodologies.

\begin{enumerate}
    \item \textbf{Enhancements in SDMap Encoding and Processing Methods.} Proper encoding and processing methods are crucial for leveraging SDMap prior information in local map perception tasks. Current studies employ relatively simple encoding and processing methods for SDMap information, whether using raster or vector representations. Future research could explore more efficient encoding and feature extraction methods.
    \item \textbf{Improvements in Aligning SDMap Prior Information with BEV Space.} Due to the accuracy limitations of GPS sensors, it is challenging to perfectly align SDMap prior information with the current BEV operational space. This spatial misalignment can affect the model's detection accuracy to some extent. Enhancing spatial alignment methods can further improve model performance. Future research could consider incorporating temporal information to enhance the alignment accuracy between SDMap prior information and BEV space.
    \item \textbf{Inference of Road Topological Relationships.} The topological relationships in local map can be divided into two branches: the topological relationships between roads (primarily representing road connectivity) and the topological relationships between roads and traffic signs (including traffic control signals and other directional signs). Enhancing scene understanding of the road environment is crucial for high-level intelligent driving tasks. The OpenLane-v2 dataset\cite{Openlane-v2} is the first public dataset providing topological relationships between roads and between roads and traffic signs. Current research focusing on this area is still limited. Future work could model the topological structures of road networks and the scene understanding tasks of traffic signs using graph neural network models.
    \item \textbf{Incorporating More SDMap Prior Information.} Existing research has demonstrated that incorporating more road-type information can enhance model performance. However, beyond the basic road network positions and road types, an SDMap can provide richer prior information. For example, OpenStreetMap\cite{Openstreetmap} offers additional information such as the number of lanes, lane directions, and road topological relationships. Future research could attempt to integrate this diverse information as SDMap priors to enhance the robustness and accuracy of local map perception models.
\end{enumerate}

\section{Conclusion}  \label{sec:conclusion}
\noindent In this article, the literature on local map construction using SDMap was reviewed, highlighting the pivotal role of SDMap in this task. The definition and core aspects of local map construction with SDMap were presented, demonstrating their significance in developing accurate and reliable maps. Commonly used public datasets and their corresponding evaluation metrics were enumerated.

This study reviews the core processes of state-of-the-art technologies, concentrating on data representation and encoding methods for sensor data, including lidar, camera, and radar inputs. It also examines sophisticated sensor fusion strategies for consolidating data from diverse sensors, evaluating their comparative advantages and constraints.

The evaluation prospects and design trends of local map construction models were discussed. This included addressing emerging challenges, such as improving SDMap alignment with BEV perspectives and enhancing encoding and processing methods. The potential of incorporating detailed SDMap prior information to model road topological relationships was considered, with the goal of improving scene understanding and supporting higher-level intelligent driving tasks.

{\small
\bibliographystyle{unsrt}
\bibliography{TITS}

\begin{thebibliography}{10}

\bibitem{VLMCGTex}
Xin Chen and Lei Yu.
\newblock Visual localization and map construction based on ground texture.
\newblock In Yingmin Jia, Weicun Zhang, Yongling Fu, and Jiqiang Wang, editors, {\em Proceedings of 2023 Chinese Intelligent Systems Conference}, pages 465--478, Singapore, 2023. Springer Nature Singapore.

\bibitem{SMERF}
Katie~Z Luo, Xinshuo Weng, Yan Wang, Shuang Wu, Jie Li, Kilian~Q Weinberger, Yue Wang, and Marco Pavone.
\newblock Augmenting lane perception and topology understanding with standard definition navigation maps.
\newblock {\em arXiv preprint arXiv:2311.04079}, 2023.

\bibitem{OMR}
Dongkwon Jin and Chang-Su Kim.
\newblock Omr: Occlusion-aware memory-based refinement for video lane detection, 2024.

\bibitem{RVLD}
Dongkwon Jin, Dahyun Kim, and Chang-Su Kim.
\newblock Recursive video lane detection, 2023.

\bibitem{Laneaf}
Hala Abualsaud, Sean Liu, David~B Lu, Kenny Situ, Akshay Rangesh, and Mohan~M Trivedi.
\newblock Laneaf: Robust multi-lane detection with affinity fields.
\newblock {\em IEEE Robotics and Automation Letters}, 6(4):7477--7484, 2021.

\bibitem{LaneATT}
Lucas Tabelini, Rodrigo Berriel, Thiago~M Paixao, Claudine Badue, Alberto~F De~Souza, and Thiago Oliveira-Santos.
\newblock Keep your eyes on the lane: Real-time attention-guided lane detection.
\newblock In {\em Proceedings of the IEEE/CVF conference on computer vision and pattern recognition}, pages 294--302, 2021.

\bibitem{Streammapnet}
Tianyuan Yuan, Yicheng Liu, Yue Wang, Yilun Wang, and Hang Zhao.
\newblock Streammapnet: Streaming mapping network for vectorized online hd map construction.
\newblock In {\em Proceedings of the IEEE/CVF Winter Conference on Applications of Computer Vision}, pages 7356--7365, 2024.

\bibitem{Openstreetmap}
Mordechai Haklay and Patrick Weber.
\newblock Openstreetmap: User-generated street maps.
\newblock {\em IEEE Pervasive computing}, 7(4):12--18, 2008.

\bibitem{kitti}
Andreas Geiger, Philip Lenz, Christoph Stiller, and Raquel Urtasun.
\newblock Vision meets robotics: The kitti dataset.
\newblock {\em The International Journal of Robotics Research}, 32(11):1231--1237, 2013.

\bibitem{nuscenes}
Holger Caesar, Varun Bankiti, Alex~H Lang, Sourabh Vora, Venice~Erin Liong, Qiang Xu, Anush Krishnan, Yu~Pan, Giancarlo Baldan, and Oscar Beijbom.
\newblock nuscenes: A multimodal dataset for autonomous driving.
\newblock In {\em Proceedings of the IEEE/CVF conference on computer vision and pattern recognition}, pages 11621--11631, 2020.

\bibitem{apolloscape}
Xinyu Huang, Xinjing Cheng, Qichuan Geng, Binbin Cao, Dingfu Zhou, Peng Wang, Yuanqing Lin, and Ruigang Yang.
\newblock The apolloscape dataset for autonomous driving.
\newblock In {\em Proceedings of the IEEE conference on computer vision and pattern recognition workshops}, pages 954--960, 2018.

\bibitem{Argoverse}
Ming-Fang Chang, John Lambert, Patsorn Sangkloy, Jagjeet Singh, Slawomir Bak, Andrew Hartnett, De~Wang, Peter Carr, Simon Lucey, Deva Ramanan, et~al.
\newblock Argoverse: 3d tracking and forecasting with rich maps.
\newblock In {\em Proceedings of the IEEE/CVF conference on computer vision and pattern recognition}, pages 8748--8757, 2019.

\bibitem{Openlane}
Ahmed Ghazy and Mohamed Shalan.
\newblock Openlane: The open-source digital asic implementation flow.
\newblock In {\em Proc. Workshop on Open-Source EDA Technol.(WOSET)}, 2020.

\bibitem{Waymo}
Pei Sun, Henrik Kretzschmar, Xerxes Dotiwalla, Aurelien Chouard, Vijaysai Patnaik, Paul Tsui, James Guo, Yin Zhou, Yuning Chai, Benjamin Caine, et~al.
\newblock Scalability in perception for autonomous driving: Waymo open dataset.
\newblock In {\em Proceedings of the IEEE/CVF conference on computer vision and pattern recognition}, pages 2446--2454, 2020.

\bibitem{Curvelane}
Hang Xu, Shaoju Wang, Xinyue Cai, Wei Zhang, Xiaodan Liang, and Zhenguo Li.
\newblock Curvelane-nas: Unifying lane-sensitive architecture search and adaptive point blending.
\newblock In {\em Computer Vision--ECCV 2020: 16th European Conference, Glasgow, UK, August 23--28, 2020, Proceedings, Part XV 16}, pages 689--704. Springer, 2020.

\bibitem{Once-3dlane}
Fan Yan, Ming Nie, Xinyue Cai, Jianhua Han, Hang Xu, Zhen Yang, Chaoqiang Ye, Yanwei Fu, Michael~Bi Mi, and Li~Zhang.
\newblock Once-3dlanes: Building monocular 3d lane detection.
\newblock In {\em Proceedings of the IEEE/CVF Conference on Computer Vision and Pattern Recognition}, pages 17143--17152, 2022.

\bibitem{Openlane-v2}
Huijie Wang, Tianyu Li, Yang Li, Li~Chen, Chonghao Sima, Zhenbo Liu, Bangjun Wang, Peijin Jia, Yuting Wang, Shengyin Jiang, et~al.
\newblock Openlane-v2: A topology reasoning benchmark for unified 3d hd mapping.
\newblock {\em Advances in Neural Information Processing Systems}, 36, 2024.

\bibitem{Bevformer}
Zhiqi Li, Wenhai Wang, Hongyang Li, Enze Xie, Chonghao Sima, Tong Lu, Yu~Qiao, and Jifeng Dai.
\newblock Bevformer: Learning bird’s-eye-view representation from multi-camera images via spatiotemporal transformers.
\newblock In {\em European conference on computer vision}, pages 1--18. Springer, 2022.

\bibitem{lss}
Jonah Philion and Sanja Fidler.
\newblock Lift, splat, shoot: Encoding images from arbitrary camera rigs by implicitly unprojecting to 3d.
\newblock In {\em Computer Vision--ECCV 2020: 16th European Conference, Glasgow, UK, August 23--28, 2020, Proceedings, Part XIV 16}, pages 194--210. Springer, 2020.

\bibitem{ResNet}
Kaiming He, Xiangyu Zhang, Shaoqing Ren, and Jian Sun.
\newblock Deep residual learning for image recognition.
\newblock In {\em Proceedings of the IEEE conference on computer vision and pattern recognition}, pages 770--778, 2016.

\bibitem{Mobilenets}
Andrew~G Howard, Menglong Zhu, Bo~Chen, Dmitry Kalenichenko, Weijun Wang, Tobias Weyand, Marco Andreetto, and Hartwig Adam.
\newblock Mobilenets: Efficient convolutional neural networks for mobile vision applications.
\newblock {\em arXiv preprint arXiv:1704.04861}, 2017.

\bibitem{efficientnet}
Mingxing Tan and Quoc Le.
\newblock Efficientnet: Rethinking model scaling for convolutional neural networks.
\newblock In {\em International conference on machine learning}, pages 6105--6114. PMLR, 2019.

\bibitem{vov}
Youngwan Lee, Joong-won Hwang, Sangrok Lee, Yuseok Bae, and Jongyoul Park.
\newblock An energy and gpu-computation efficient backbone network for real-time object detection.
\newblock In {\em Proceedings of the IEEE/CVF conference on computer vision and pattern recognition workshops}, pages 0--0, 2019.

\bibitem{FPN}
Tsung-Yi Lin, Piotr Doll{\'a}r, Ross Girshick, Kaiming He, Bharath Hariharan, and Serge Belongie.
\newblock Feature pyramid networks for object detection.
\newblock In {\em Proceedings of the IEEE conference on computer vision and pattern recognition}, pages 2117--2125, 2017.

\bibitem{vit}
Alexey Dosovitskiy, Lucas Beyer, Alexander Kolesnikov, Dirk Weissenborn, Xiaohua Zhai, Thomas Unterthiner, Mostafa Dehghani, Matthias Minderer, Georg Heigold, Sylvain Gelly, et~al.
\newblock An image is worth 16x16 words: Transformers for image recognition at scale.
\newblock {\em arXiv preprint arXiv:2010.11929}, 2020.

\bibitem{Swin}
Ze~Liu, Yutong Lin, Yue Cao, Han Hu, Yixuan Wei, Zheng Zhang, Stephen Lin, and Baining Guo.
\newblock Swin transformer: Hierarchical vision transformer using shifted windows.
\newblock In {\em Proceedings of the IEEE/CVF international conference on computer vision}, pages 10012--10022, 2021.

\bibitem{eva}
Yuxin Fang, Quan Sun, Xinggang Wang, Tiejun Huang, Xinlong Wang, and Yue Cao.
\newblock Eva-02: A visual representation for neon genesis.
\newblock {\em arXiv preprint arXiv:2303.11331}, 2023.

\bibitem{bert}
Jacob Devlin, Ming-Wei Chang, Kenton Lee, and Kristina Toutanova.
\newblock Bert: Pre-training of deep bidirectional transformers for language understanding.
\newblock {\em arXiv preprint arXiv:1810.04805}, 2018.

\bibitem{mae}
Kaiming He, Xinlei Chen, Saining Xie, Yanghao Li, Piotr Doll{\'a}r, and Ross Girshick.
\newblock Masked autoencoders are scalable vision learners.
\newblock In {\em Proceedings of the IEEE/CVF conference on computer vision and pattern recognition}, pages 16000--16009, 2022.

\bibitem{mim}
Zhenda Xie, Zheng Zhang, Yue Cao, Yutong Lin, Yixuan Wei, Qi~Dai, and Han Hu.
\newblock On data scaling in masked image modeling.
\newblock In {\em Proceedings of the IEEE/CVF Conference on Computer Vision and Pattern Recognition}, pages 10365--10374, 2023.

\bibitem{P-MapNet}
Zhou Jiang, Zhenxin Zhu, Pengfei Li, Huan-ang Gao, Tianyuan Yuan, Yongliang Shi, Hang Zhao, and Hao Zhao.
\newblock P-mapnet: Far-seeing map generator enhanced by both sdmap and hdmap priors.
\newblock {\em arXiv preprint arXiv:2403.10521}, 2024.

\bibitem{Pointnet}
Charles~R Qi, Hao Su, Kaichun Mo, and Leonidas~J Guibas.
\newblock Pointnet: Deep learning on point sets for 3d classification and segmentation.
\newblock In {\em Proceedings of the IEEE conference on computer vision and pattern recognition}, pages 652--660, 2017.

\bibitem{Maplite2.0}
Teddy Ort, Jeffrey~M Walls, Steven~A Parkison, Igor Gilitschenski, and Daniela Rus.
\newblock Maplite 2.0: Online hd map inference using a prior sd map.
\newblock {\em IEEE Robotics and Automation Letters}, 7(3):8355--8362, 2022.

\bibitem{maplite}
Teddy Ort, Jeffrey~M Walls, Steven~A Parkison, Igor Gilitschenski, and Daniela Rus.
\newblock Maplite 2.0: Online hd map inference using a prior sd map.
\newblock {\em IEEE Robotics and Automation Letters}, 7(3):8355--8362, 2022.

\bibitem{Priorlane}
Qibo Qiu, Haiming Gao, Wei Hua, Gang Huang, and Xiaofei He.
\newblock Priorlane: A prior knowledge enhanced lane detection approach based on transformer.
\newblock In {\em 2023 IEEE International Conference on Robotics and Automation (ICRA)}, pages 5618--5624. IEEE, 2023.

\bibitem{MapVision}
Zhongyu Yang, Mai Liu, Jinluo Xie, Yueming Zhang, Chen Shen, Wei Shao, Jichao Jiao, Tengfei Xing, Runbo Hu, and Pengfei Xu.
\newblock Mapvision: Cvpr 2024 autonomous grand challenge mapless driving tech report.
\newblock {\em arXiv preprint arXiv:2406.10125}, 2024.

\bibitem{MapTR}
Bencheng Liao, Shaoyu Chen, Xinggang Wang, Tianheng Cheng, Qian Zhang, Wenyu Liu, and Chang Huang.
\newblock Maptr: Structured modeling and learning for online vectorized hd map construction.
\newblock {\em arXiv preprint arXiv:2208.14437}, 2022.

\bibitem{MapEX}
R{\'e}my Sun, Li~Yang, Diane Lingrand, and Fr{\'e}d{\'e}ric Precioso.
\newblock Mind the map! accounting for existing map information when estimating online hdmaps from sensor data.
\newblock {\em arXiv preprint arXiv:2311.10517}, 2023.

\bibitem{3d-lanenet}
Noa Garnett, Rafi Cohen, Tomer Pe'er, Roee Lahav, and Dan Levi.
\newblock 3d-lanenet: end-to-end 3d multiple lane detection.
\newblock In {\em Proceedings of the IEEE/CVF International Conference on Computer Vision}, pages 2921--2930, 2019.

\bibitem{Gen-lanenet}
Yuliang Guo, Guang Chen, Peitao Zhao, Weide Zhang, Jinghao Miao, Jingao Wang, and Tae~Eun Choe.
\newblock Gen-lanenet: A generalized and scalable approach for 3d lane detection.
\newblock In {\em Computer Vision--ECCV 2020: 16th European Conference, Glasgow, UK, August 23--28, 2020, Proceedings, Part XXI 16}, pages 666--681. Springer, 2020.

\bibitem{Persformer}
Li~Chen, Chonghao Sima, Yang Li, Zehan Zheng, Jiajie Xu, Xiangwei Geng, Hongyang Li, Conghui He, Jianping Shi, Yu~Qiao, et~al.
\newblock Persformer: 3d lane detection via perspective transformer and the openlane benchmark.
\newblock In {\em European Conference on Computer Vision}, pages 550--567. Springer, 2022.

\bibitem{Anchor3dlane}
Shaofei Huang, Zhenwei Shen, Zehao Huang, Zi-han Ding, Jiao Dai, Jizhong Han, Naiyan Wang, and Si~Liu.
\newblock Anchor3dlane: Learning to regress 3d anchors for monocular 3d lane detection.
\newblock In {\em Proceedings of the IEEE/CVF Conference on Computer Vision and Pattern Recognition}, pages 17451--17460, 2023.

\bibitem{Bev-lanedet}
Ruihao Wang, Jian Qin, Kaiying Li, Yaochen Li, Dong Cao, and Jintao Xu.
\newblock Bev-lanedet: An efficient 3d lane detection based on virtual camera via key-points.
\newblock In {\em Proceedings of the IEEE/CVF Conference on Computer Vision and Pattern Recognition}, pages 1002--1011, 2023.

\bibitem{BeMapNet}
Limeng Qiao, Wenjie Ding, Xi~Qiu, and Chi Zhang.
\newblock End-to-end vectorized hd-map construction with piecewise bezier curve.
\newblock In {\em Proceedings of the IEEE/CVF Conference on Computer Vision and Pattern Recognition}, pages 13218--13228, 2023.

\bibitem{Curveformer}
Yifeng Bai, Zhirong Chen, Zhangjie Fu, Lang Peng, Pengpeng Liang, and Erkang Cheng.
\newblock Curveformer: 3d lane detection by curve propagation with curve queries and attention.
\newblock In {\em 2023 IEEE International Conference on Robotics and Automation (ICRA)}, pages 7062--7068. IEEE, 2023.

\bibitem{Grouplane}
Zhuoling Li, Chunrui Han, Zheng Ge, Jinrong Yang, En~Yu, Haoqian Wang, Hengshuang Zhao, and Xiangyu Zhang.
\newblock Grouplane: End-to-end 3d lane detection with channel-wise grouping.
\newblock {\em arXiv preprint arXiv:2307.09472}, 2023.

\bibitem{Maptrv2}
Bencheng Liao, Shaoyu Chen, Yunchi Zhang, Bo~Jiang, Qian Zhang, Wenyu Liu, Chang Huang, and Xinggang Wang.
\newblock Maptrv2: An end-to-end framework for online vectorized hd map construction.
\newblock {\em arXiv preprint arXiv:2308.05736}, 2023.

\bibitem{Pivotnet}
Wenjie Ding, Limeng Qiao, Xi~Qiu, and Chi Zhang.
\newblock Pivotnet: Vectorized pivot learning for end-to-end hd map construction.
\newblock In {\em Proceedings of the IEEE/CVF International Conference on Computer Vision}, pages 3672--3682, 2023.

\bibitem{HIMap}
Yi~Zhou, Hui Zhang, Jiaqian Yu, Yifan Yang, Sangil Jung, Seung-In Park, and ByungIn Yoo.
\newblock Himap: Hybrid representation learning for end-to-end vectorized hd map construction.
\newblock In {\em Proceedings of the IEEE/CVF Conference on Computer Vision and Pattern Recognition}, pages 15396--15406, 2024.

\bibitem{LaneCPP}
Maximilian Pittner, Joel Janai, and Alexandru~P Condurache.
\newblock Lanecpp: Continuous 3d lane detection using physical priors.
\newblock In {\em Proceedings of the IEEE/CVF Conference on Computer Vision and Pattern Recognition}, pages 10639--10648, 2024.

\bibitem{PVALane}
Zewen Zheng, Xuemin Zhang, Yongqiang Mou, Xiang Gao, Chengxin Li, Guoheng Huang, Chi-Man Pun, and Xiaochen Yuan.
\newblock Pvalane: Prior-guided 3d lane detection with view-agnostic feature alignment.
\newblock In {\em Proceedings of the AAAI Conference on Artificial Intelligence}, volume~38, pages 7597--7604, 2024.

\bibitem{img1}
Ziye Chen, Kate Smith-Miles, Bo~Du, Guoqi Qian, and Mingming Gong.
\newblock An efficient transformer for simultaneous learning of bev and lane representations in 3d lane detection.
\newblock {\em arXiv preprint arXiv:2306.04927}, 2023.

\bibitem{mapqr}
Zihao Liu, Xiaoyu Zhang, Guangwei Liu, Ji~Zhao, and Ningyi Xu.
\newblock Leveraging enhanced queries of point sets for vectorized map construction.
\newblock In {\em European Conference on Computer Vision}, pages 461--477. Springer, 2025.

\bibitem{MapBench}
Xiaoshuai Hao, Mengchuan Wei, Yifan Yang, Haimei Zhao, Hui Zhang, Yi~Zhou, Qiang Wang, Weiming Li, Lingdong Kong, and Jing Zhang.
\newblock Is your hd map constructor reliable under sensor corruptions?
\newblock {\em arXiv preprint arXiv:2406.12214}, 2024.

\bibitem{Mask2Map}
Sehwan Choi, Jungho Kim, Hongjae Shin, and Jun~Won Choi.
\newblock Mask2map: Vectorized hd map construction using bird's eye view segmentation masks.
\newblock {\em arXiv preprint arXiv:2407.13517}, 2024.

\bibitem{img2}
Wencheng Han and Jianbing Shen.
\newblock Decoupling the curve modeling and pavement regression for lane detection.
\newblock {\em arXiv preprint arXiv:2309.10533}, 2023.

\bibitem{HDmapnet}
Qi~Li, Yue Wang, Yilun Wang, and Hang Zhao.
\newblock Hdmapnet: An online hd map construction and evaluation framework.
\newblock In {\em 2022 International Conference on Robotics and Automation (ICRA)}, pages 4628--4634. IEEE, 2022.

\bibitem{LiLaDet}
Runkai Zhao, Yuwen Heng, Yuanda Gao, Shilei Liu, Heng Wang, Changhao Yao, Jiawen Chen, and Weidong Cai.
\newblock Advancements in 3d lane detection using lidar point clouds: From data collection to model development.
\newblock {\em arXiv preprint arXiv:2309.13596}, 2023.

\bibitem{Petrv2}
Yingfei Liu, Junjie Yan, Fan Jia, Shuailin Li, Aqi Gao, Tiancai Wang, and Xiangyu Zhang.
\newblock Petrv2: A unified framework for 3d perception from multi-camera images.
\newblock In {\em Proceedings of the IEEE/CVF International Conference on Computer Vision}, pages 3262--3272, 2023.

\bibitem{Vectormapnet}
Yicheng Liu, Tianyuan Yuan, Yue Wang, Yilun Wang, and Hang Zhao.
\newblock Vectormapnet: End-to-end vectorized hd map learning.
\newblock In {\em International Conference on Machine Learning}, pages 22352--22369. PMLR, 2023.

\bibitem{fusion1}
Yasin Yen{\i}aydin and Klaus~Werner Schmidt.
\newblock Sensor fusion of a camera and 2d lidar for lane detection.
\newblock In {\em 2019 27th Signal Processing and Communications Applications Conference (SIU)}, pages 1--4. IEEE, 2019.

\bibitem{fusion2}
Nathir~A Rawashdeh, Jeremy~P Bos, and Nader~J Abu-Alrub.
\newblock Camera--lidar sensor fusion for drivable area detection in winter weather using convolutional neural networks.
\newblock {\em Optical Engineering}, 62(3):031202--031202, 2023.

\bibitem{fusion4}
Min Bai, Gellert Mattyus, Namdar Homayounfar, Shenlong Wang, Shrinidhi~Kowshika Lakshmikanth, and Raquel Urtasun.
\newblock Deep multi-sensor lane detection.
\newblock In {\em 2018 IEEE/RSJ International Conference on Intelligent Robots and Systems (IROS)}, pages 3102--3109. IEEE, 2018.

\bibitem{fusion5}
Feihu Zhang, Daniel Clarke, and Alois Knoll.
\newblock Vehicle detection based on lidar and camera fusion.
\newblock In {\em 17th International IEEE Conference on Intelligent Transportation Systems (ITSC)}, pages 1620--1625. IEEE, 2014.

\bibitem{BLOS-BEV}
Hang Wu, Zhenghao Zhang, Siyuan Lin, Tong Qin, Jin Pan, Qiang Zhao, Chunjing Xu, and Ming Yang.
\newblock Blos-bev: Navigation map enhanced lane segmentation network, beyond line of sight.
\newblock In {\em 2024 IEEE Intelligent Vehicles Symposium (IV)}, pages 3212--3219. IEEE, 2024.

\bibitem{FlexMap}
Maximilian Leitenstern, Florian Sauerbeck, Dominik Kulmer, and Johannes Betz.
\newblock Flexmap fusion: Georeferencing and automated conflation of hd\~{} maps with openstreetmap.
\newblock {\em arXiv preprint arXiv:2404.10879}, 2024.

\bibitem{Bayesian}
Yuxuan Xia, Erik Stenborg, Junsheng Fu, and Gustaf Hendeby.
\newblock Bayesian simultaneous localization and multi-lane tracking using onboard sensors and a sd map.
\newblock {\em arXiv preprint arXiv:2405.04290}, 2024.

\bibitem{TopoLogic}
Yanping Fu, Wenbin Liao, Xinyuan Liu, Yike Ma, Feng Dai, Yucheng Zhang, et~al.
\newblock Topologic: An interpretable pipeline for lane topology reasoning on driving scenes.
\newblock {\em arXiv preprint arXiv:2405.14747}, 2024.

\bibitem{LGmap}
Kuang Wu, Sulei Nian, Can Shen, Chuan Yang, and Zhanbin Li.
\newblock Lgmap: Local-to-global mapping network for online long-range vectorized hd map construction.
\newblock {\em arXiv preprint arXiv:2406.13988}, 2024.

\bibitem{lsm}
Guang Li, Jianwei Ren, Quanyun Zhou, Anbin Xiong, and Kuiyuan Yang.
\newblock Leveraging sd map to assist the openlane topology.

\bibitem{UniHDMap}
Genghua Kou, Fan Jia, Dongming Wu, Yingfei Liu, Ying Li, and Tiancai Wang.
\newblock Unihdmap: Unified lane elements detection for topology hd map construction.

\bibitem{RoadPainter}
Zhongxing Ma, Shuang Liang, Yongkun Wen, Weixin Lu, and Guowei Wan.
\newblock Roadpainter: Points are ideal navigators for topology transformer.
\newblock {\em arXiv preprint arXiv:2407.15349}, 2024.

\bibitem{Enhancing_Online}
Hengyuan Zhang, David Paz, Yuliang Guo, Arun Das, Xinyu Huang, Karsten Haug, Henrik~I Christensen, and Liu Ren.
\newblock Enhancing online road network perception and reasoning with standard definition maps.
\newblock {\em arXiv preprint arXiv:2408.01471}, 2024.

\bibitem{GKT}
Shaoyu Chen, Tianheng Cheng, Xinggang Wang, Wenming Meng, Qian Zhang, and Wenyu Liu.
\newblock Efficient and robust 2d-to-bev representation learning via geometry-guided kernel transformer.
\newblock {\em arXiv preprint arXiv:2206.04584}, 2022.

\bibitem{segformer}
Enze Xie, Wenhai Wang, Zhiding Yu, Anima Anandkumar, Jose~M Alvarez, and Ping Luo.
\newblock Segformer: Simple and efficient design for semantic segmentation with transformers.
\newblock {\em Advances in neural information processing systems}, 34:12077--12090, 2021.

\bibitem{deeplabv3}
Liang-Chieh Chen, Yukun Zhu, George Papandreou, Florian Schroff, and Hartwig Adam.
\newblock Encoder-decoder with atrous separable convolution for semantic image segmentation, 2018.

\bibitem{coco}
Tsung-Yi Lin, Michael Maire, Serge Belongie, Lubomir Bourdev, Ross Girshick, James Hays, Pietro Perona, Deva Ramanan, C.~Lawrence Zitnick, and Piotr Dollár.
\newblock Microsoft coco: Common objects in context, 2015.

\bibitem{Lanesegnet}
Tianyu Li, Peijin Jia, Bangjun Wang, Li~Chen, Kun Jiang, Junchi Yan, and Hongyang Li.
\newblock Lanesegnet: Map learning with lane segment perception for autonomous driving.
\newblock {\em arXiv preprint arXiv:2312.16108}, 2023.

\bibitem{YOLOv8}
Rejin Varghese and Sambath. M.
\newblock Yolov8: A novel object detection algorithm with enhanced performance and robustness.
\newblock {\em 2024 International Conference on Advances in Data Engineering and Intelligent Computing Systems (ADICS)}, pages 1--6, 2024.

\end{thebibliography}
}
\newpage

\begin{IEEEbiography}[{\includegraphics[width=1in,height=1.25in,clip,keepaspectratio]{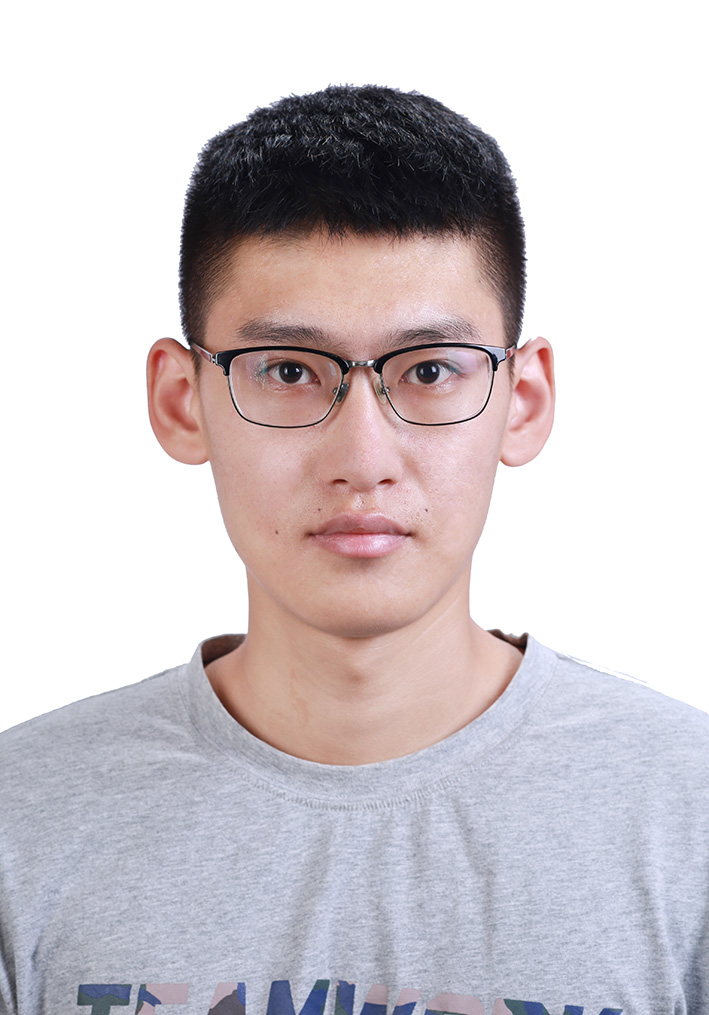}}]{Jiaqi Li} received the B.S. degree in Surveying and Mapping Engineering from the Qinghai University, Xining, China, in 2022. He is currently pursuing a M.S. degree at the Department of Civil Engineering, Tsinghua University, Beijing, China. Li’s research interests include computer vision and online HDMapping.\end{IEEEbiography}

\begin{IEEEbiography}[{\includegraphics[width=1in,height=1.25in,clip,keepaspectratio]{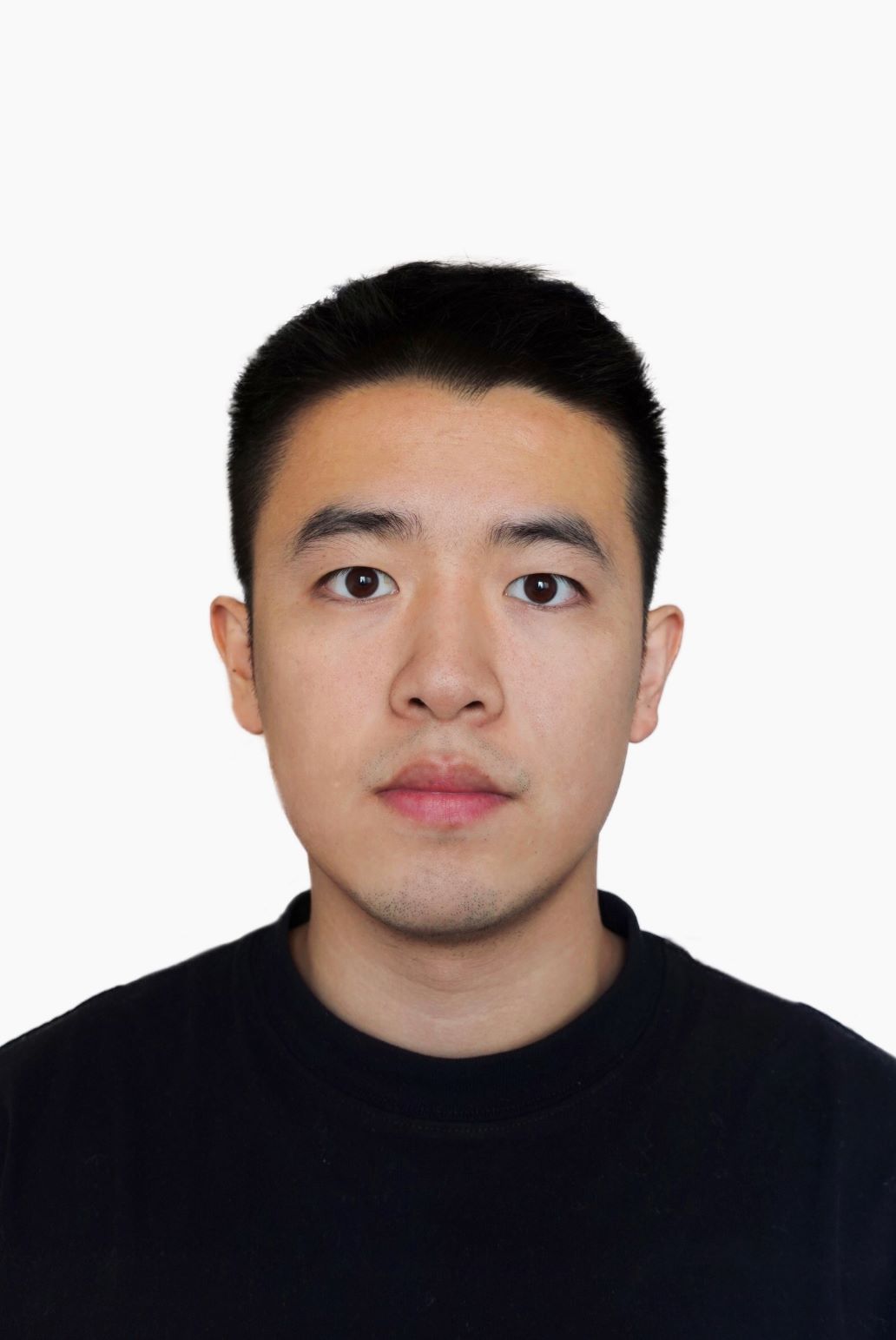}}]{Pingfan Jia} is currently pursuing a M.S. degree from the School of Computer Science at Beihang University in 2022. He is currently working at the Multimodal Perception and Computing Research Lab, where he is engaged in cutting-edge Computer Vision research. His research interests are focused on intelligent driving Perception and 3D Semantic Scene Completion.\end{IEEEbiography}

\begin{IEEEbiography}[{\includegraphics[width=1in,height=1.25in,clip,keepaspectratio]{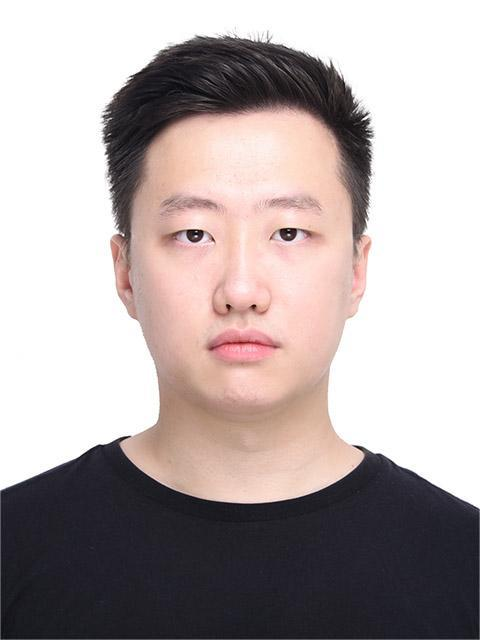}}]{Jiaxing Chen} received an M.S. degree in Electrical and Computer Engineering from the University of Illinois, Chicago, USA, in 2021 and worked as an Algorithm engineer at the National Innovation Center of Intelligent and Connected Vehicles, Beijing, China from 2021 to 2022. He is currently pursuing a Ph.D. degree at the School of Vehicle and Mobility, Tsinghua University, Beijing, China. Chen’s research interests include computer vision and perception based on Vehicle-Road-Cloud architecture.\end{IEEEbiography}

\begin{IEEEbiography}[{\includegraphics[width=1in,height=1.25in,clip,keepaspectratio]{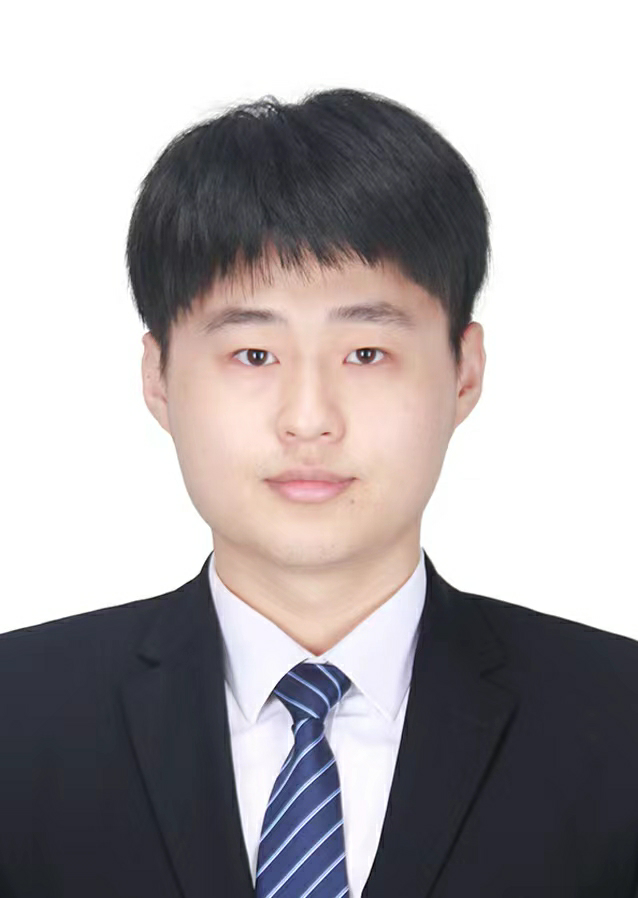}}]{Jiaxi Liu} received a B.E. and an M.E.degree in Mechanical Engineering at the School of Vehicle and Mobility, Tsinghua University. He is now pursuing a Ph.D. degree in the Civil and Environment Engineering Department, at the University of Wisconsin-Madison. His research interests include collaborative perception, real-time perception, vehicle-road-cloud integration systems, and LLM-assist intelligent driving.\end{IEEEbiography}

\newpage
\begin{IEEEbiography}[{\includegraphics[width=1in,height=1.25in,clip,keepaspectratio]{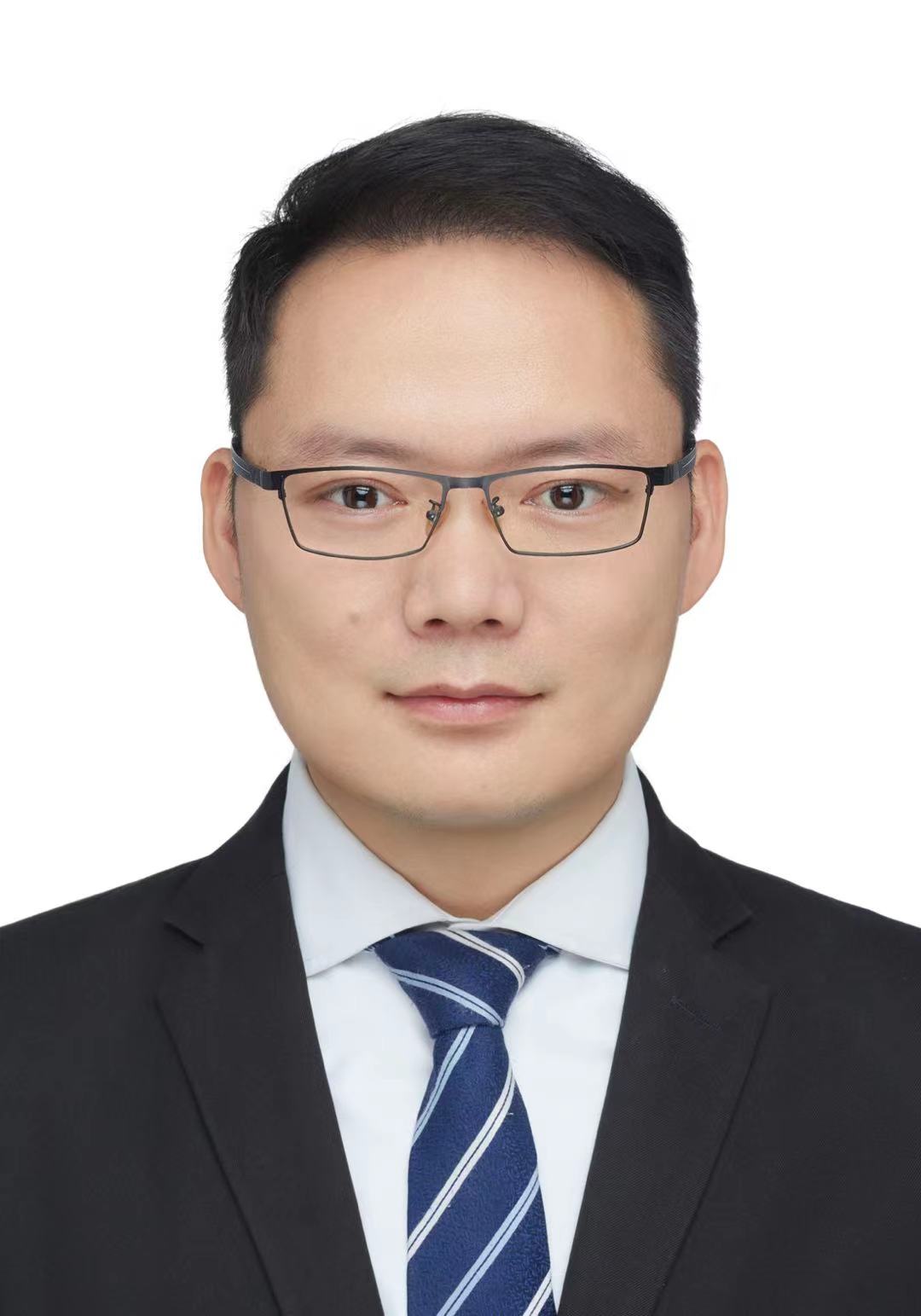}}]{Lei He} received his B.S. in Beijing University of Aeronautics and Astronautics, China, in 2013, and the Ph.D. in the National Laboratory of Pattern Recognition, Chinese Academy of Sciences, in 2018. From then to 2021, Dr. He served as a postdoctoral fellow in the Department of Automation, Tsinghua University, Beijing, China. He worked as the research leader of the intelligent driving algorithm at Baidu and NIO from 2018 to 2023. He is a Research Scientist in automotive engineering with Tsinghua University. His research interests include Perception, SLAM, Planning, and Control.\end{IEEEbiography}

\vspace{-11.0cm}
\begin{IEEEbiography}[{\includegraphics[width=1in,height=1.25in,clip,keepaspectratio]{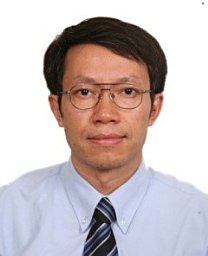}}]{Keqiang Li} received the B.Tech. degree from the Tsinghua University of China, Beijing, China, in 1985, and the M.S. and Ph.D. degrees in mechanical engineering from the Chongqing University of China, Chongqing, China, in 1988 and 1995, respectively. He is currently a Professor at the School of Vehicle and Mobility, Tsinghua University. His main research interests include automotive control systems, driver assistance system, and networked dynamics and control. He is leading the national key project on Intelligent and Connected Vehicles, China. He has authored more than 200 papers and is a co-inventor of more than 80 patents in China and Japan. Dr. Li was a Fellow Member of the Society of Automotive Engineers of China, the Chairperson of Expert Committee of the China Industrial Technology Innovation Strategic Alliance for ICVs (CAICV), and CTO of China ICV Research Institute Company Ltd. (CICV). He served on the Editorial Boards of the International Journal of Vehicle Autonomous Systems. He was the recipient of Changjiang Scholar Program Professor, National Award for Technological Invention in China.\end{IEEEbiography}

\end{CJK}
\end{document}